\documentclass[lettersize,journal]{IEEEtran}
\usepackage{amsmath,amsfonts}
\usepackage{algorithmic}
\usepackage{algorithm}
\usepackage{array}
\usepackage[caption=false,font=normalsize,labelfont=sf,textfont=sf]{subfig}
\usepackage{textcomp}
\usepackage{stfloats}
\usepackage{url}
\usepackage{verbatim}
\usepackage{graphicx}
\usepackage{cite}
\usepackage{todonotes}
\usepackage{color}
\usepackage{hyperref}

\usepackage{amsmath}
\DeclareMathOperator*{\argmax}{argmax} 

\newcolumntype{P}[1]{>{\centering\arraybackslash}p{#1}}
\newcolumntype{C}[1]{>{\centering\arraybackslash}m{#1}}

\def \textHT [#1]{\color{red}$\mathbf{#1}$\color{black}}

\def \textLT
[#1]{\color{blue}$\mathbf{#1}$\color{black}}

\begin{document}
\title{iSimLoc: Visual Global Localization for Previously Unseen Environments with Simulated Images}

\author{Peng Yin,~\IEEEmembership{Member,~IEEE,},
        Ivan Cisneros,
        Ji Zhang, 
        Howie Choset, ~\IEEEmembership{Fellow,~IEEE,},
        and Sebastian Scherer, ~\IEEEmembership{Senior Member,~IEEE,}
\thanks{Peng Yin, Ivan Cisneros, Ji Zhang, Howie Choset, and Sebastian Scherer are with the Robotics Institute, Carnegie Mellon University, Pittsburgh, PA 15213, USA. {(pyin2, icisnero, zhangji, choset, basti)@andrew.cmu.edu}.}
\thanks{Corresponding author: Peng Yin (pyin2@andrew.cmu.edu)}
}

\markboth{IEEE Transactions on Robotics (T-RO). 
Preprint Version. January, 2022}
{Yin \MakeLowercase{\textit{et al.}}: iSimLoc: Hybrid Visual Localization in Non-visited Environment for Large-scale Aerial Navigation}


\maketitle

\begin{abstract}
The visual camera is an attractive device in beyond visual line of sight (B-VLOS) drone operation, since they are low in size, weight, power, and cost, and can provide redundant modality to GPS failures. However, state-of-the-art visual localization algorithms are unable to match visual data that have a significantly different appearance due to illuminations or viewpoints. This paper presents iSimLoc, a condition/viewpoint consistent hierarchical global re-localization approach. The place features of iSimLoc can be utilized to search target images under changing appearances and viewpoints. Additionally, our hierarchical global re-localization module refines in a coarse-to-fine manner, allowing iSimLoc to perform a fast and accurate estimation. We evaluate our method on one dataset with appearance variations and one dataset that focuses on demonstrating large-scale matching over a long flight in complicated environments. On our two datasets, iSimLoc achieves 88.7\% and 83.8\% successful retrieval rates with 1.5s inferencing time, compared to 45.8\% and 39.7\% using the next best method. These results demonstrate robust localization in a range of environments.
\end{abstract}


\section{Introduction}
\label{sec:introduction}

    \IEEEPARstart{U}{nmanned} Aerial Vehicles (UAVs) have become popular in different non-military and commercial applications such as cargo transport~\cite{Auto:larson2019autonomous}, surveillance~\cite{Intro:land_mapping}, precision agriculture~\cite{Intro:precision_agriculture} and  search-rescue tasks~\cite{Intro:search_rescue1,Intro:search_rescue2}.
    Current UAVs primarily rely on GPS as their only source for global position information, making their localization systems fragile to GPS outages, an issue which is often addressed with safety pilots~\cite{Intro:uav_landing}. 
    However, in the future, beyond visual line of sight (B-VLOS) flight will require a backup source of global position to achieve sufficient reliability.
    
\begin{figure}[t]
    	\centering
        \includegraphics[width=\linewidth]{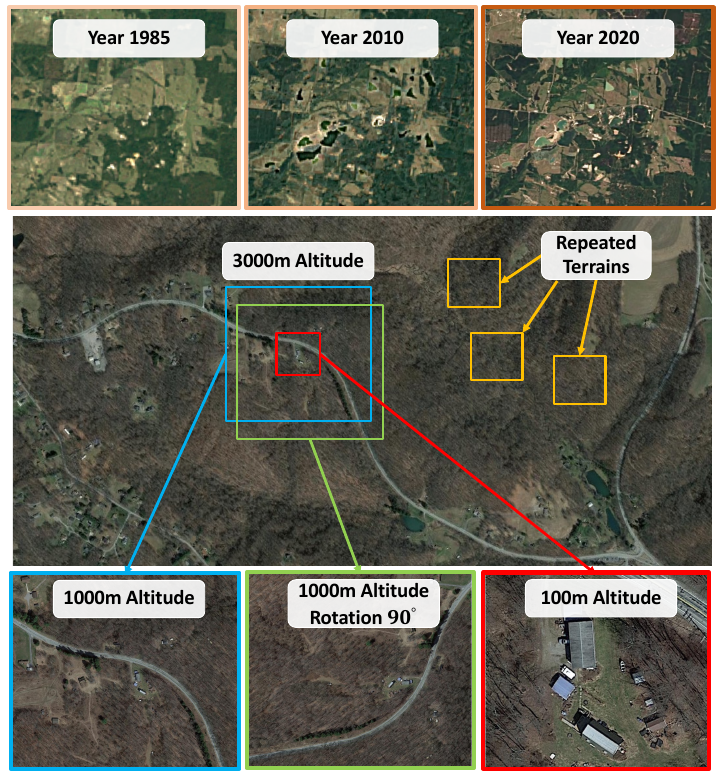}
    	\caption{\textbf{Challenges for Visual Terrain Navigation.}
        Visual terrain navigation systems will encounter (a) varying appearance over time (shown in first row), (b) varying viewpoints (i.e. orientation and altitude shown in last row), and (c) repeated terrain (shown in middle figure).}
    	\label{fig:altitude_diff}
    \end{figure}

    To this end we develop an alternate method that is redundant with GPS, and is based on the idea of Visual Terrain-Relative Navigation (VTRN)~\cite{SR:VTRN}.
    Cameras are more attractive for UAVs than active sensors such as LiDAR since cameras are passive, able to run over long distances, have a low SWaP-C (Size, Weight, Power, and Cost), and have a large field-view.
    As shown in Fig.~\ref{fig:altitude_diff}, challenges of VTRN for localization include:
    \begin{itemize}
        \item \textbf{Appearance Changes:}  Appearances of a particular area may change drastically under different illumination and weather conditions, making data association challenging.
        \item \textbf{Viewpoint Differences:} When re-visiting the same place, the UAV cannot guarantee that it will revisit the exact same position, orientation, and altitude, which requires the method to be robust to variations.
        \item \textbf{High-similarity at High Altitudes:} At high altitudes, there is a high likelihood of areas with repeated and homogeneous geometries, such as forests and flat ground, leading to false data associations.
    \end{itemize}
    \color{black}
    
    We propose a novel VTRN approach, iSimLoc, which is invariant to external condition changes, and provides robust \textit{Global Re-localization} in large-scale terrain/urban areas. 
    iSimLoc trains a place recognition model for non-visited environments by leveraging overhead imagery. As it is based on our previous work~\cite{Yin:SphereVLAD,Yin:i3dLoc}, iSimLoc is also invariant to appearance changes caused by illumination and viewpoint differences.
    When re-localizing within a large area, iSimLoc relies on a coarse-to-fine localization method to balance efficiency and accuracy. 
    The contributions of iSimLoc are:

    \begin{itemize}
        \item 
        \textbf{Real-to-Sim Conditional-domain Transfer:} A \textit{Conditional Domain Transfer Module} (CDTM) to transform the raw images into a constant simulated domain. The \textit{CDTM} extracts both geometric and conditional features from raw images for conditional invariant place recognition.
        Our experimental results on urban and terrain areas demonstrate that the conditional domain transfer module improves recognition ability for non-visited areas.
        \item 
        \textbf{Viewpoint-invariant Place Recognition:} Most VTRN systems depend on fixed viewpoints between test/reference queries. iSimLoc, on the other hand, calculates viewpoint-invariant descriptors and estimates relative orientation through usage of the \textit{Pose Estimation Module} (PEM).
        Specifically, the \textit{PEM} utilizes spherical harmonic features, which are orientation-equivariant for the same location.
        \item 
        \textbf{Hierarchical Localization System:}
        For global re-localization, iSimLoc matches hierarchically starting with a coarse estimate at a high altitude and refining with repeated cropping of images to make more accurate estimates. This method balances efficiency and accuracy, which is essential for re-localization in large maps.
    \end{itemize}
    
    In our experiment results, we present an extensive evaluation of our system on two unique datasets: 
    (1) The \textit{CMU Campus} dataset consists of $160$ trajectories taken under different lighting conditions targeting an urban environment, which was collected using a quadcopter flying on Carnegie Mellon University's campus; 
    (2) and the \textit{Large Terrain} dataset consists of one $150$km trajectory covering both urban and natural terrain, which was collected using a helicopter.
    On both datasets, iSimLoc outperforms all other relative place recognition methods and achieves a $88.7\%$ successful recognition rate for urban areas and $83.8\%$ for natural terrain.
    Leveraging overhead images helps our system attain higher generalization ability for unseen environments.
    We include a discussion section and a conclusion section to analyze the advantages and shortcomings of the current iSimLoc approach, as well as to examine potential future work.

\section{Related Works}
\label{sec:related_works}
    Visual geo-localization is defined as finding geographic coordinates (and possibly camera orientation) for a given query image.
    Based on survey~\cite{survey:geo-vis}, visual geo-localization is divided into city-scale and natural localization.
    The main challenges are environmental condition changes, viewpoint differences, and large-scale re-localization.

    To deal with changes in environmental conditions, Mishkin \emph{et al.}~\cite{Condition:mishkin2015place} modified BoW approach~\cite{FeatureCapturer:BoW2} with multiple descriptors and adaptive thresholds to better cope with large-scale changes in environments.
    Bhavit~\emph{et al.}~\cite{Intro:ge_uav1} introduce a visual terrain navigation method where reference images are rendered from Google Earth (GE) satellite images. However, because such overhead images from GE are captured several years prior to test time, visual differences between reference/testing streams reduce localization accuracy.
    To deal with this issue, Mollie~\emph{et al.}~\cite{Intro:ge_uav2} utilized an Autoencoder network to transfer raw images into overhead images, ignoring local dynamic differences or environmental condition changes.
    However, this method cannot handle illumination, weather, and seasonal changes, reducing generalization ability for unseen conditions.
    In their recent work~\cite{ScienceRobotic:Season}, Anthony~\emph{et al.} provide a seasonally invariant deep transform neural network to convert seasonal images into a stable and invariant domain for visual terrain navigation.
    This method targets high-altitude flying modes as depicted in Fig.~\ref{fig:altitude_diff}, where there exist rich unchanging geometric features that persist even in different seasons. However, in lower altitudes, this method's transfer-ability will be negatively affected by occlusions of 3D objects and highly variable lighting conditions from different times of the day.
    Another solution to deal with disturbances from environmental conditions is to match horizontal lines extracted from a query image against those rendered from digital elevation models (DEM)~\cite{Intro:DEM}.
    In~\cite{Intro:HorizonMountain}, Baatz~\emph{et al.} demonstrate terrain localization by leveraging this method.
    Similarly, Bertil~\emph{et al.}~\cite{Intro:HorizonNet} introduced an accurate camera localization for unmanned surface vessels (USVs) by aligning horizontal lines with coastal structures.
    A significant drawback of horizontal line-based approaches is dependence on rich 3D geometric structures, such as mountainous or coastal areas.
    Thus, performance will be reduced in homogeneous and plain environments.
    Similar to~\cite{ScienceRobotic:Season}, our iSimLoc method transfers raw images into a constant domain.

    \begin{table*} [htbp]
        \caption{Properties of different visual terrain relative navigation methods.}
        \centering
        \begin{tabular}{|C{1.7cm}|C{2cm}|C{1.5cm}|C{1.5cm}|C{1.5cm}|C{1.5cm}|C{1.5cm}|C{1.5cm}|}
            \hline
            \textbf{Method}
            & \textbf{Class}
            & \textbf{Environment}
            & \textbf{Condition}
            & \textbf{Viewpoint}
            & \textbf{Altitude}
            & \textbf{Localization Accuracy}
            & \textbf{Global Localization}
            \\
            \hline
            Anthony~\emph{et al.}~\cite{ScienceRobotic:Season} & Domain Transfer & Natural & Changing & Known & High & $\leq50m$ & No\\
            \hline
            Patel~\emph{et al.}~\cite{Intro:ge_uav1} & Domain Transfer & Hybrid & Changing & Unknown & $\leq50m$ & $\leq10m$ & No\\
            \hline
            Bianchi~\emph{et al.}~\cite{Intro:ge_uav2} & Domain Transfer & Hybrid & Changing & Known & Low & $\leq10m$ & No\\
            \hline
            Baatz~\emph{et al.}~\cite{Intro:HorizonMountain} & Geometric & Mountain & Fixed & Unknown & High & N/A & No\\
            \hline
            Jeong-Kyun~\emph{et al.}~\cite{Intro:VanishPoint} & Geometric & City & Changing & Unknown & Low & N/A & No\\
            \hline
            Baatz~\emph{et al.}~\cite{Intro:DEM} & Geometric & Mountain & Fixed & Unknown & High & N/A & Yes\\
            \hline
            Pluckter~\emph{et al.}~\cite{Landing:precision_landing} & Geometric & Natural & Fixed & Unknown & Low & 1m & No\\
            \hline
            Arandjelovic~\emph{et al.}~\cite{PR:netvlad} & Place Recognition & City & Fixed & Unknown & Low & N/A & No\\
            \hline
            Michael~\emph{et al.}~\cite{VPR:SeqSLAM} & Place Recognition & Hybrid & Changing & Known & Low & N/A & No\\
            \hline
            Peng~\emph{et al.}~\cite{i3dloc} & Place Recognition & Hybrid & Changing & Unknown & Low & 1m & No\\ \hline
            iSimLoc & Place Recognition & Hybrid & Changing & Unknown & Hybrid & 20m & Yes\\
            \hline
        \end{tabular}
        \label{tab:CorrelationAveraging}
    \end{table*}

    Viewpoint difference poses another significant challenge for accurate localization.
    As depicted in Fig.~\ref{fig:altitude_diff}, orientation and altitude differences can significantly change the appearance of a location from the original perspective.
    In both~\cite{ScienceRobotic:Season} and~\cite{Intro:HorizonMountain}, the authors conducted image alignment over very high altitude flights ($7,000\sim 15,000m$).
    In such cases, top-down images are able to capture rich distinguishable geometric features for accurate matching.
    However, not all applications can leverage this benefit since the Federal Aviation Administration (FAA) sets the flying altitude limit for UAV drones to around $120m$. Helicopters usually fly at $\approx300m$ when at lower altitudes, encountering changing viewpoints and environmental conditions. Additionally, current visual terrain navigation methods usually assume that relative orientations between raw inputs of drones/helicopter and reference images are small. Thus, these methods focus on position estimation and do not include robust relative orientation estimation. In real applications, a noisy GPS signal may result in a sizeable initial orientation estimation error, resulting in image alignment failures and, consequently, a loss of accuracy. Yet, most image-based alignment methods are unable to estimate corresponding orientations~\cite{Intro:ge_uav1,Intro:ge_uav2,ScienceRobotic:Season}. Jeong-Kyun~\emph{et al.}~\cite{Intro:VanishPoint} introduced a vanishing point-based camera orientation estimation method. This method is suitable for urban indoor and outdoor scenes, as detection of vanishing points is based on line segments.
    Shichao~\emph{et al.}~\cite{Intro:cubeslam} proposed CubeSLAM for camera pose estimation and localization based on extracted cubic objects.
    Both line features and cubic objects can be relatively easily captured in urban environments; however, they are sparsely present in natural terrain.
    Baatz~\emph{et al.}~\cite{Intro:DEM} used semantic information (tree, river) as constraints for camera pose estimation. However, a significant limitation comes from unreliability of detecting semantic objects.
    When using visual terrain navigation at lower altitudes, image distortion caused by viewpoint changes will further reduce feature stability.
    To extract viewpoint-invariant descriptors in both low and high altitudes --- similar to our previous 3D place recognition work~\cite{Max:spherevlad, i3dloc} --- iSimLoc utilizes spherical harmonics~\cite{Sphere:SO3_learning} to extract orientation-equivariant features from spherical perspectives.
    iSimLoc also estimates the relative orientation between test and reference based on constant amplitude of spherical harmonics.
    This ability further improves online localization robustness even for long-term visual terrain navigation.

    Most existing VTRN methods focus on local re-localization against a reference image and share the assumption that the robot has relatively good estimates of its position and orientation.  
    However, in real-world applications, environmental conditions and viewpoints (both orientation and altitude) may change dramatically and simultaneously.
    Additionally, similar and repeated geometric features in natural terrain environments, such as in forests and flat plain ground, will further reduce localization success rates.
    Most VTRN methods can, thus, hardly deal with global re-localization in large-scale environments without a great deal of assistance from GPS.
    
    In Table.~\ref{tab:CorrelationAveraging}, we compare different properties of current VTRN methods.
    Domain transfer-based methods are mainly designed to deal with changing environmental conditions. 
    Notably different to~\cite{ScienceRobotic:Season}, they ignore viewpoint differences, making them most suitable for local re-localization, whereas iSimLoc also includes viewpoint-invariant feature extraction to handle viewpoint differences.
    On other hand, most geometric-based methods mainly target changing viewpoints under constant environmental assumptions. 
    Few methods consider both conditional and viewpoint differences in visual terrain localization at the same time.
    In contrast to our previous work \textit{i3dLoc}~\cite{i3dloc} on \textit{Robotics: Science and Systems 2021}, we use a place descriptor for locally re-localization, iSimLoc aims to provide robust visual global re-localization for large-scale environments by leveraging overhead imagery.
    
    \begin{figure*}[t]
    	\centering
        \includegraphics[width=\linewidth]{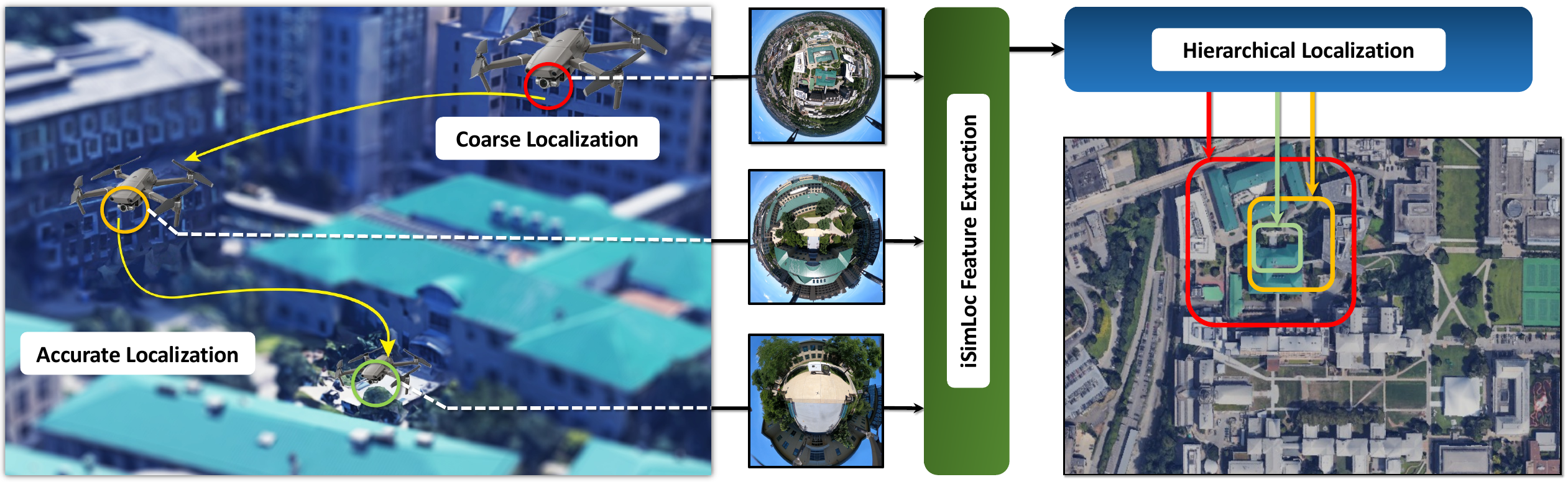}
    	\caption{\textbf{iSimLoc system framework.}
        For high and low altitudes, iSimLoc extracts a condition-(illumination) and viewpoint-invariant place descriptor. Only the descriptor needs to be stored and matched. Larger field of views help iSimLoc to provide an initial guess, while narrower field of view perspectives provide rich local geometry features for accurate localization. iSimLoc matches hierarchically, which enables us to balance search efficiency and accuracy.
        }
    	\label{fig:pipeline}
    \end{figure*}
    
    \section{System Overview}
    \label{sec:system_oveview}
    iSimLoc uses publicly available overhead imagery and camera images from a UAV as input to provide condition- and viewpoint- invariant global re-localization in large-scale terrain/urban environments. The iSimLoc framework consists of the following three steps:
    1) \textit{Data Collection} to provide paired simulation/raw data to train the conditional-domain transfer module; 
    2) \textit{Place Feature Extraction} for viewpoint-invariant place recognition, and 
    3) \textit{Global Re-localization} for hierarchical robust localization.

    \subsection{Data Collection}
    \label{sec:Data_collection}
    We use two data collection platforms to generate paired simulated and real-world data for training and evaluation.
    The first dataset is recorded under changing conditions and at different altitudes around the Carnegie Mellon University campus.
    The recording platform is a quadcopter with a mounted omnidirectional camera. We name it ``\textit{CMU Campus}'' dataset.
    The second dataset (``\textit{Large Terrain}'') is a flight from Cambridge, Ohio, to Pittsburgh, Pennsylvania using a helicopter with a downward-facing pinhole camera.
    
    For the \textit{CMU Campus} dataset, we use way-point following mode to make repeated passes along fixed trajectories to collect data on the same path but under different conditions (illumination, weather, time, etc.). 
    \textit{Large Terrain} dataset includes $150$km trajectory.
    We generate a paired sim-to-real dataset for both platforms by exporting the trajectories' GPS information and collecting publicly available nadir overhead imagery from Google Earth.
    Due to low frequency of the GPS data, we interpolate data among two GPS points to generate time-synced sim-to-real paired images.
    In Section~\ref{sec:experiments}, we expand further on the details of our data generation procedure.
    
    \subsection{Feature Extraction}
    \label{sec:feature_extraction}
    Due to condition and viewpoint differences, extracting invariant place descriptors is the most critical factor in visual localization. 
    In iSimLoc, we first use a conditional domain transfer learning module (CDTM) to convert raw images into a constant geometric domain.
    As depicted in Fig.~\ref{fig:pipeline}, the CDTM is forced to extract conditional and geometric features with an orthogonal relationship distribution.
    With extracted geometric features, iSimLoc learns viewpoint invariant place descriptors with the Pose Extraction Module (PEM).
    The PEM can estimate place descriptors' similarities while ignoring their orientation differences, and predict the relative yaw difference given matched descriptors for a particular area.
    Since spherical viewpoints will reduce geometric differences under varying altitudes, the PEM is also robust to local altitude differences.
    Due to these invariance properties, the iSimLoc place feature model is able to recognize places under different viewpoints, which increases sampling efficiency for global re-localization.

    \subsection{Global Re-localization}
    \label{sec:online_tracking}
    Without GPS assistance, current visual terrain relative navigation methods may quickly encounter tracking failures in repeated and homogeneous terrain areas, especially for long-term navigation tasks.
    Our global re-localization procedure uses extracted iSimLoc place features and a particle filter~\cite{Yin:Multi_resolution} in a coarse-to-fine hierarchical refinement method to overcome the limitations of other methods.
    Higher altitude images provide a coarse position estimation over a large search area, reducing the number of initial particles needed.
    At lower altitudes, visual inputs capture more geometric structures, which produces more accurate final matches.
    Since features are orientation invariant, particles only need to be sampled within Euclidean space $\mathcal{R}^3$, instead of $\mathcal{SO}(3)$ space, significantly reducing the total number of particles used.

    \section{Our Approach}
    \label{sec:OA}
    
    \begin{figure*}[!th]
        \centering
        \includegraphics[width=\linewidth]{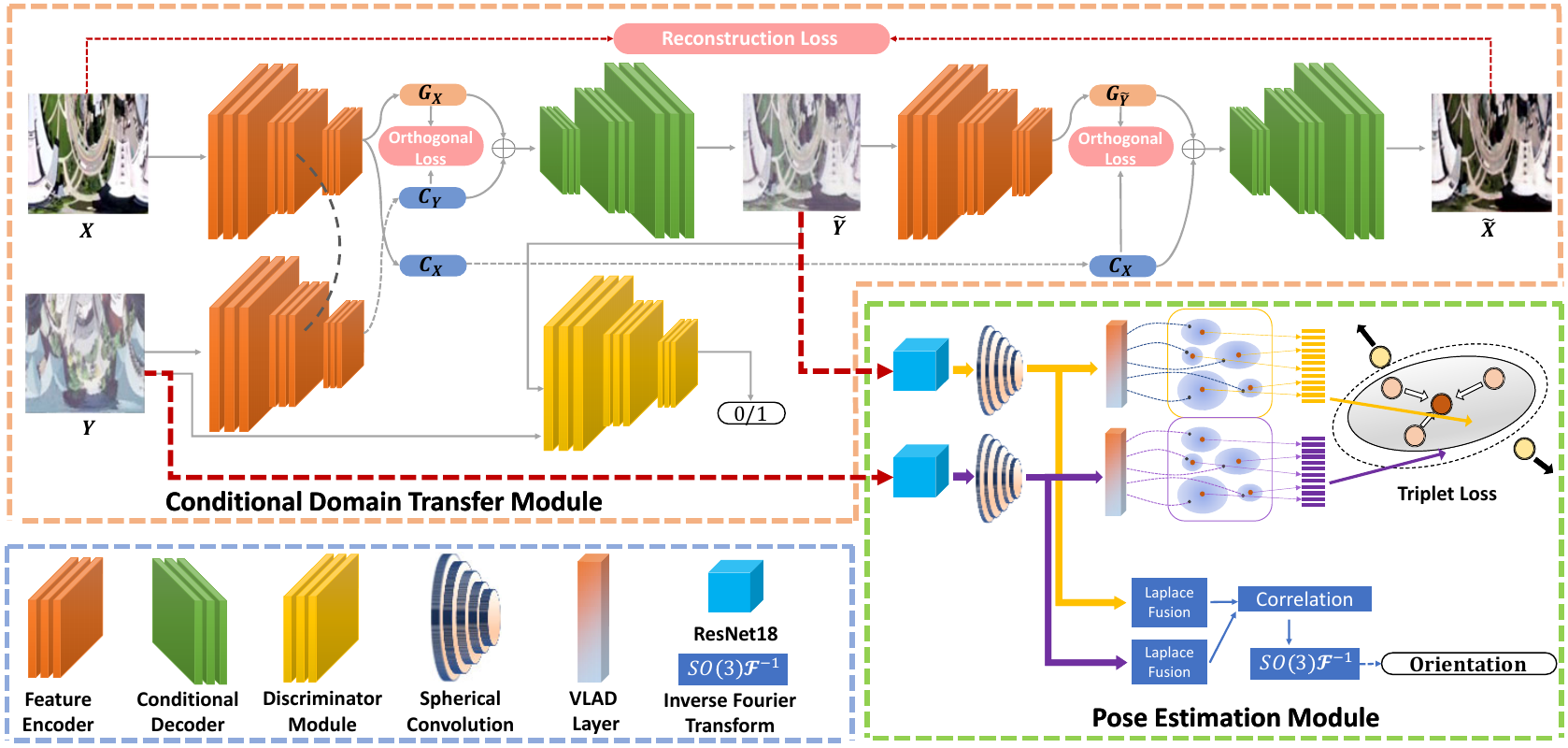}
        \caption{\textbf{The network structure of iSimLoc.}
        iSimLoc consists of a conditional domain transfer module (CDTM) to transform raw visual inputs into overhead images, where environmental conditions are applied to improve generalization ability for unseen environments, and a pose estimation module (PEM) to simultaneously estimate viewpoint-invariant place features and relative orientations.
        }
        \label{fig:framework}
    \end{figure*}
    
    As illustrated in Fig.~\ref{fig:framework}, the main idea of iSimLoc is to learn corresponding conditional- and viewpoint- invariant place descriptors that help to localize against overhead imagery. It consists of three modules:
    \begin{enumerate}
        \item A conditional domain transfer module to convert raw images into a constant overhead image domain.
        \item A pose extraction module to recognize the place and orientation based on estimated features' similarities.
        \item A hierarchical global re-localization module to track position at both low and high altitudes.
    \end{enumerate}
 
    \subsection{Conditional Domain Transfer Module}
    \label{sec:CDTM}
    To generate constant geometric features from visual inputs under different environmental conditions, we construct a conditional domain transfer module (\textit{CDTM}), which includes a feature encoder, a conditional decoder, and a discriminator.

Before introducing the feature extraction process, we analyze the relationship of information entropy to viewpoints.
    Naturally, raw visual images include both geometric features $Z_G$, which depend on 3D geometric structures, and condition features $Z_C$, which encode an image's appearance that is caused by the combination of environmental conditions (illumination, weather, and seasons).
    As depicted in Fig.~\ref{fig:framework}, given a raw image, our encoder module  extracts both geometric features $Z_G$ and conditional features $Z_C$ simultaneously.

    Given an image $x$, $H(Z_{G}, Z_{C}|x)$ and $I(Z_{G};Z_{C}|x)$ represents joint entropy and mutual entropy.
    $H(Z_{G}|Z_{C},x)$ and $H(Z_{C}|Z_{G},x)$ are conditional entropies based on $Z_{C}$ and $Z_{G}$, respectively.
    The target of place recognition is to extract condition-invariant geometric features, related to conditional information $H(Z_G|Z_C,x)$; and restrict extracted feature's uncertainty given same image $x$, which is relative to joint entropy $H(Z_G,Z_C|x)$.
    To learn condition-invariant place descriptors, we focus on:
    \begin{itemize}
        \item Increasing condition entropy $H(Z_G|Z_C, x)$ to \textit{enrich} geometric feature $Z_G$ extraction from raw image $x$, which is independent of conditional feature $Z_C$, and
        \item Reducing joint entropy $H(Z_G,Z_C|x)$, which measures joint features' ($\{Z_G, Z_C\}$) differences when revisiting the same place.
    \end{itemize}
    
    Directly improving conditional entropy $H(Z_G|Z_C, x)$ in place recognition tasks is intractable, since each unique area may drastically vary in appearance due to different combinations of environmental conditions (illumination, weather, seasons), and it is hard to access all potential $Z_G\rightarrow Z_C$ pairs.
    Alternatively, using the information theory perspective, $H(Z_G|Z_C, x)$ can be converted into,
    \begin{align}
        H(Z_{G}|Z_{C},x) = H(Z_{G}|x)-I(Z_{G};Z_{C}|x)\label{eq:mutual_entropy}
    \end{align}
    where $H(Z_G|x)$ measures diversity of geometric features $Z_G$ based on observation $x$. This is important because in place recognition tasks, higher variance in place features provide better distinguish-ability.
    We design a generative adversarial network (GAN)~\cite{CNN:GAN} to enhance diversity of $H(Z_G|x)$, where corresponding loss metric $\mathcal{L}_{GAN}$ is written as,
    \begin{align}
        &\mathcal{L}_{GAN} 
        = \min_{\theta, \phi}\max_{\beta}E(\log(D_{\beta}(y)) +\label{ep:loss_gan} \\
        &E_{\{Z_G,Z_C\}\sim p_{\theta}(Z|x),\hat{y}\sim q_{\phi}(y|Z_G, \hat{Z}_C)}(\log(1-D_{\beta}(\hat{y})))\nonumber
    \end{align}
    where $y$ and $\hat{y}$ are real and generated overhead images, $D_{\beta}$ is the discriminator to distinguish $y$ and $\hat{y}$, $P_{\theta}$ is the encoder and $Q_{\phi}$ is the decoder, and $\beta, \theta, \phi$ are the learnable parameters.
    As provided by Goodfellow \emph{et.al}~\cite{CNN:GAN}, with iterative updating of generative network (decoder $Q_{\phi}$) and discriminator modules, an adversarial network is able to push distribution of $\hat{y}$ towards target distribution $y$.
    On the other hand, since mutual entropy $I(Z_G;Z_C|x)$ measures overlaps between geometric features $Z_G$ and condition features $Z_C$, reducing $I(Z_G;Z_C|x)$ indicates the minimum projection from $Z_C$ onto $Z_G$.
    We apply an orthogonal loss metric $\mathcal{L}_{Orth}$ to enhance the orthogonal relationship between features $Z_G$ and $Z_C$, 
    \begin{align}
        \mathcal{L}_{Orth} = 1 - \frac{Z_G\cdot Z_C}{\|Z_G\|_2 \cdot\|Z_C\|_2}
        \label{eq:loss_orth}
    \end{align}
    Conditional entropy $H(Z_G|Z_C,x)$ is able to be increased by the combination of $\mathcal{L}_{Orth}+\mathcal{L}_{GAN}$.

    To reduce joint entropy $H(Z_{C},Z_{G}|x)$, similar to CycleGAN~\cite{CycleGAN2017}, we construct an L1 loss metric between raw image $x$ and reconstructed image $\hat{x}$ as demonstrated in Fig.~\ref{fig:framework}.
    Using raw image $x$, iSimLoc generates overhead image $\hat{y}$ with geometric feature $Z_{G_{x}}$ and condition feature $Z_{C_{y}}$; then using $\hat{y}$, iSimLoc reconstructs $\hat{x}$ with $Z_{G_{\hat{y}}}$ and $Z_{C_{x}}$.
    In our previous work~\cite{i3dloc}, we prove that decreasing $H(Z_{G}, Z_{C}|x)$ corresponds to reducing image reconstruction uncertainty given sample data $x$.
    We formulate the reconstruction loss as,
    \begin{align}
        \mathcal{L}_{Recon} = &H_{\{Z_{G_x}, Z_{C_x}\}\sim P_{\theta}(x), \{Z_{G_{\hat{y}}}, Z_{C_{\hat{y}}}\}
        \sim P_{\theta}(\hat{y})} \label{ep:loss_recon} \\
        &[\log(Q_{\phi}(Z_{G_{\hat{y}}}, Z_{C_x})|x)]  \nonumber
    \end{align}
    The original $H(Z_G, Z_C|x)$ is transformed into its upper bound $\mathcal{L}_{Recon}(\hat{x}, x)$.
    Based on Eq.~\ref{ep:loss_gan}, \ref{eq:loss_orth} and \ref{ep:loss_recon}, we construct the loss metric for the conditional domain transfer module as,
    \begin{align}
        \mathcal{L}_{CDTM} = \mathcal{L}_{GAN}+\mathcal{L}_{Orth}+\mathcal{L}_{Recon}
        \label{eq:loss_cdtm}
    \end{align}

    \subsection{Pose Estimation Module}
    \label{sec:PEM}
    In iSimLoc, the Pose Estimation Module (PEM) is designed to predict viewpoint-invariant place descriptors and extract relative orientations.
    The PEM module is combined with a pre-trained ResNet18 model from torchvision\footnote{https://pytorch.org/vision/stable/models.html} for deep feature extraction, and a spherical feature-based viewpoint-invariant descriptor extractor and orientation estimator.
 
     \begin{figure}[t]
        \centering
        \includegraphics[width=\linewidth]{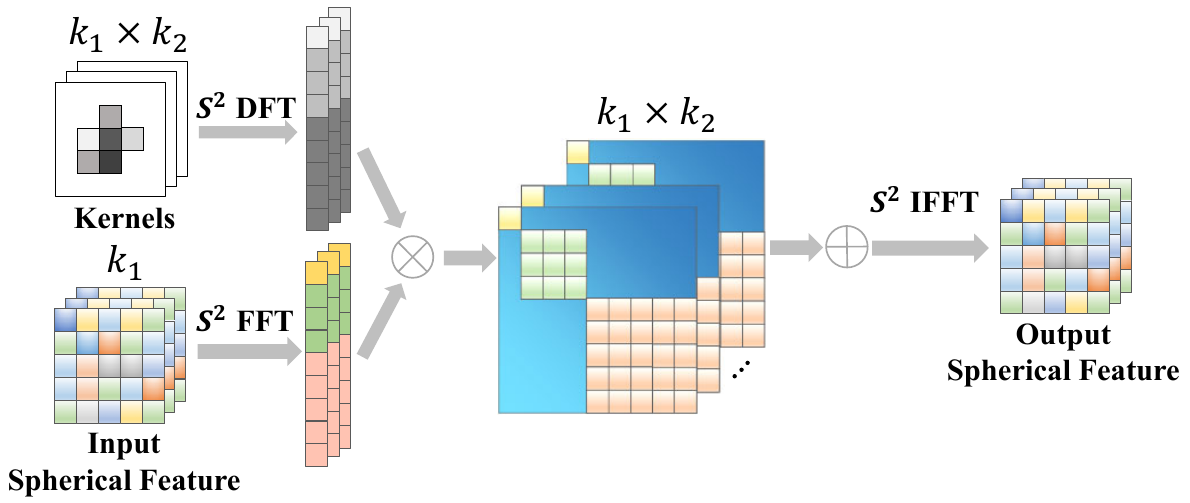}
        \caption{\textbf{Spherical convolution module.}
       Given a spherical feature $f$ and a kernel signal $h$, we first transform them into harmonic domain ($H_f$, $H_h$) using a Fast Fourier transform (FFT) and a Discrete Fourier Transform (DFT).}
        \label{fig:sphere_projection}
    \end{figure}

    \bigskip
    \subsubsection{Viewpoint-invariant Descriptor}
    \label{sec:vid}
    Using spherical (360$^{\circ}$) images is natural for viewpoint-invariant place recognition.
    However, traditional convolutional neural networks are not well suited to use with spherical images, since angular resolution is not uniform across these types of images.
    Instead of a traditional convolution, we apply the spherical convolution, which utilizes spherical harmonics present in spherical projection images.
    Spherical convolutions avoid space-varying distortions in Euclidean space by convolving spherical signals in the harmonic domain.
    The mathematical model of the spherical convolution into harmonic domain shows that it is orientation-equivariant.
    Spherical convolution of $SO(3)$ signals $f$ and $h$ ($f, h$ are functions: $SO(3) \rightarrow \mathbb{R}^K$) in rotation group $SO(3)$ is defined as,
    \begin{align}
        [f \star_{SO(3)} h](\mathbf{R}) = & \int_{SO(3)}f(\mathbf{R}^{-1}\mathbf{Q})h(\mathbf{Q})d\mathbf{Q}
    \end{align}
    where $\mathbf{R, Q} \in SO(3)$. 
    As the proof in~\cite{cohen2018spherical} shows, spherical convolutions are orientation-equivariant,
    \begin{align}
        [f \star_{SO(3)} [L_{Q}h](\mathbf{R})
        = & [L_{\mathbf{Q}}[f \star_{SO(3)} h]](\mathbf{R})
    \end{align}
    where $L_{\mathbf{Q}} (\mathbf{Q} \in SO(3))$ is a rotation operator for spherical signals.
    Convolution of two spherical signals $f$ and $h$ in the harmonics domain is computed in three steps:
    We first expand $f$ and $h$ to their spherical harmonic basis $H_f$ and $H_h$, then compute the point-wise product of harmonic coefficients, and finally invert the spherical harmonic expansion.
    In our previous works~\cite{Yin:SphereVLAD,i3dloc}, we have utilized this approach to extract orientation-invariant place descriptors for either 3D point clouds or spherical images.
    For more details on spherical harmonic properties, we suggest the reader refers to the original work in~\cite{Sphere:SO3_learning}.
    To leverage viewpoint-invariant feature extraction, we utilize VLAD networks~\cite{PR:netvlad} to cluster local orientation-equivariant features into global place descriptors.
    With the assistance of the \textit{CDTM} module, iSimLoc is able to learn condition- and viewpoint- invariant place descriptors.

    \bigskip
    \subsubsection{Orientation Estimation}
    \label{sec:OE}
    As shown in the \textit{PEM} module of Fig.~\ref{fig:framework}, given extracted spherical harmonic features from two relative spherical images, we apply an estimation module to directly obtain relative orientations.
    As shown in~\cite{Sphere:phaser}, the spherical correlation $\hat{C}$ between two spherical signals $f_1$ and $f_2$ is their inner product $\hat{C}=\langle f_1, f_2\rangle$.
    If $f_2$ is the rotated version of $f_1$, relative orientation $r\in SO(3)$ can be estimated by maximizing $\hat{C}$,
    \begin{align}
        \argmax_{r\in SO(3)} \langle f_1,r^{-1}f_2 \rangle
    \end{align}
    Based on the orthogonal property of the spherical harmonics and the magnitudes property of harmonic signals~\cite{Sphere:FFT}, the above equation can be evaluated using the spherical Fourier coefficients.
    Since this part is beyond the scope of our paper, please refer to~\cite{Sphere:phaser} for a more detailed derivation.

    \begin{figure}[t]
    	\centering
        \includegraphics[width=\linewidth]{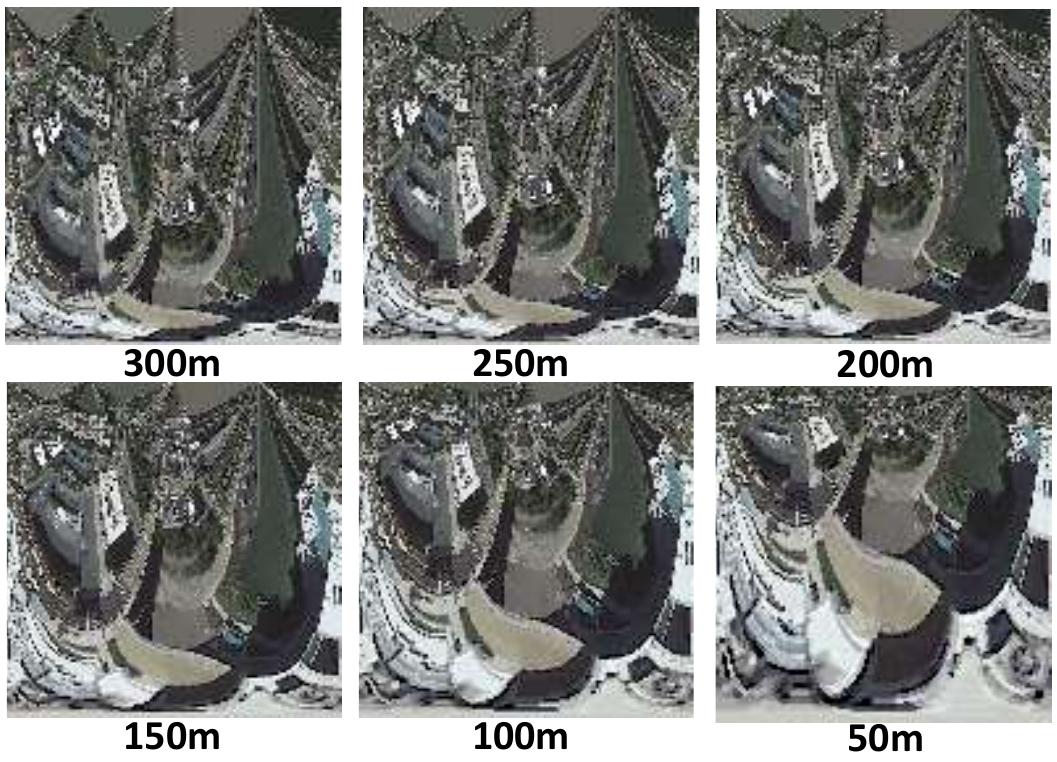}
    	\caption{\textbf{Spherical Images at Varying Altitudes.}
        Overhead imagery projected into spherical images at different altitudes.
        Higher altitudes capture more context, while lower altitudes capture richer local details. 
        }
    	\label{fig:hiloc_inputs}
    \end{figure}

    \subsubsection{Pose Loss Metric}
    \label{sec:pose_loss}
    To enable end-to-end training for viewpoint-invariant place feature extraction, we apply individual triplet-loss in both raw image domain and simulation (i.e. overhead) image domain separately, and also cross triplet-loss between the two domains.
    To illustrate loss functions, we first describe necessary definitions.
    In both real-world and simulated image domain, for each query image $M_k$ we provide the following tuples $\mathcal{T}_k = [S_k, \{S_{rot}\}_k, \{S_{pos}\}_k, \{S_{neg}\}_k]$.
    $S_k$ is encoded spherical place descriptors from $M_k$, $\{S_{rot}\}_k$ is a set of descriptors from the same position of $M_k$ but for different orientations.
    As was done in relative place recognition work~\cite{PR:netvlad}, we also provide positive $\{S_{pos}\}_k$ and negative $\{S_{neg}\}_k$ features based on distance to $M_k$.
    We construct paired tuples within raw image domain $\mathcal{S}^{V}$ and simulation image domain $\mathcal{S}^{S}$.
    Ideally, for each scenery, the place descriptor should be invariant to orientation and sensitive to translation differences, thus we design the following loss function,
    \begin{align}
        &L_{Individual}(\mathcal{T}_k) = \\
        &\max_{i\in \{S_{pos}\}_k, j\in\{S_{neg}\}_k}([\lambda_1 + d(S_k, S_{pos_i}) -
        d(S_k, S_{neg_i})]_+) + \nonumber \\
        &\max_{i,j,k}([\lambda_2 + d(S_{rot_j}, S_{pos_i}) -d(S_{rot_j}, S_{neg_i})]_+) \nonumber
    \end{align}
    where $d(\cdot)$ denotes Euclidean distance, and $[.]_+$ denotes the hinge loss.
    $\lambda_1$ and $\lambda_2$ are constant thresholds to control the margin between feature differences of different Euclidean distances.
    Meanwhile, we also define a domain learning metric to reduce cross-domain feature differences:
    \begin{align}
        &L_{CrossDomain}(\mathcal{T}_k) = \\
        &\max_{i\in \{S_{pos}\}_k, j\in\{S_{neg}\}_k}([\lambda_3 + d(S_k^{V}, S_{pos_i}^{S})-
        d(S_k^{V}, S_{neg_i}^{S})]_+) + \nonumber
    \end{align}
    where $\lambda_3$ is constant threshold to control margin between raw and simulated image features.
    By combining both individual metrics and cross domain metric, we obtain the following loss function for the Pose Estimation Module,
    \begin{align}
        \mathcal{L}_{PEM} = \mathcal{L}_{Individual}^V + \mathcal{L}_{Individual}^S + L_{CrossDomain}
        \label{ep:joint}
    \end{align}
    In our application, $\lambda_1$, $\lambda_2$ are set to $0.5$ and $\lambda_3$ is set to $1.0$.
    During the training procedure, we first train the domain transfer module with paired real-world and simulated images; then we use the pre-trained transfer model to produce conditional- and viewpoint-invariant place descriptors for use in training the rest of the network.
    
    \subsection{Hierarchical Localization}
    \label{sec:hiloc}
    Visual ambiguity is unavoidable during high-altitude flying in large-scale terrain relative navigation. Therefore, we devise a hierarchical localization module that helps iSimLoc achieve robust localization with a coarse-to-fine searching strategy.
    In contrast with place features used in other works~\cite{Intro:ge_uav1,Intro:ge_uav2}, our place descriptor is symmetric to viewpoint differences, and so each area only requires one descriptor at a certain resolution.
    
    As shown in Fig.~\ref{fig:hiloc_inputs}, given one test image at full resolution, we generate spherical projections ($[256\times 256]$) at different altitudes.
    Higher altitude images capture larger context for coarse global localization, which will reduce the number of initial particles.
    Given a potential search area of size $[M_1\times M_2]m^2$, the potential field-of-view of each particle is based on altitude $H$.
    We define the active searching radius as $H\tan{45}^\circ=H$.
    At the lowest resolution level, particles are sampled uniformly on a reference overhead image, and we define a ratio $R_{olp}\in [0,1]$ to control the overlaps between two sampled areas.
    The initial number of particles $P_{init}$ is decided by
    \begin{align}
        P_{init}=\frac{M_1\cdot M_2}{(H\cdot(1-R_{olp}))^2} \label{eq:initial_paritcle}
    \end{align}
    where a higher ratio $R_{olp}$ results in more initial particles.

    The weight of each particle $\omega_i$ is estimated by computing the cosine similarity,
    \begin{align}
        \omega_{i} = \cos(S_{i}, S_{c}) = \frac{S_{i}\cdot S_{c}}{\|S_{i}\| \cdot \|S_{c}\|}
    \end{align}
    where $S_{c}$ is the encoded feature from aerial vehicle image, $S_{i}$ is the feature of $i$-th particle.
    New particles are re-generated from the highest weighted particles, with random translations added.
    The distribution of particles converge to a local area through iterative updates.
    We use weights to estimate particles' convergence $N_{eff}=1/(\sum \left(\omega_{k}^{i}\right)^2)$, and determine whether to change resolution level.
    
    When particles converge into potential areas, we decrease altitude and remove $10\sim 30\%$ of particles, then new particles are re-sampled around the remaining original weighted particles.
    By iteratively updating particles, our method accurately matches while reducing computational burden.
    Given a $[M_1\times M_2]$ reference overhead image, searching for the best match through brute-force directly at the lowest altitude $H_1$ has a complexity of $O\left(P_{init}\right)\cdot O\left(C \right)$.
    Bound $O\left(C \right)$ includes reference image generation, feature extraction and feature similarity calculation, and is a constant for different altitudes.
    Assume the highest altitude is $\alpha$ times $H_1$, then we calculate computation complexity between \textit{Brute-force} and \textit{Hierarchical} searching by,
        \begin{align}
        \mathcal{C}_{\frac{Brute-Force}{Hierarchical}}&=\frac{\frac{O\left(M_1\cdot M_2\right)\cdot O\left(C \right)}{(H\cdot (1-R_{olp}))^2}}{O\left(\sum_{i=0}^{l_{max}}\frac{M_1\cdot M_2}{(\alpha H\cdot (1-R_{olp}))^2}\cdot 0.8^i\right)\cdot O\left(C \right)} \\ \nonumber 
        &=O\left(\frac{\alpha^2\cdot (1-0.2)}{1-0.8^{l_{max}}}\right)
    \end{align}
    where $l_{max}$ is the maximum number of layers of hierarchical searching.
    When we let $\alpha=3$ and $l_{max}=5$, we expect \textit{hierarchical} approach to be around $2.7$ times faster than the \textit{Brute-force} method.
    Besides increased matching efficiency, our hierarchical search method also increases robustness to orientation and altitude variations.

\section{Experiments}
\label{sec:experiments}

\begin{figure}[t]
    	\centering
        \includegraphics[width=\linewidth]{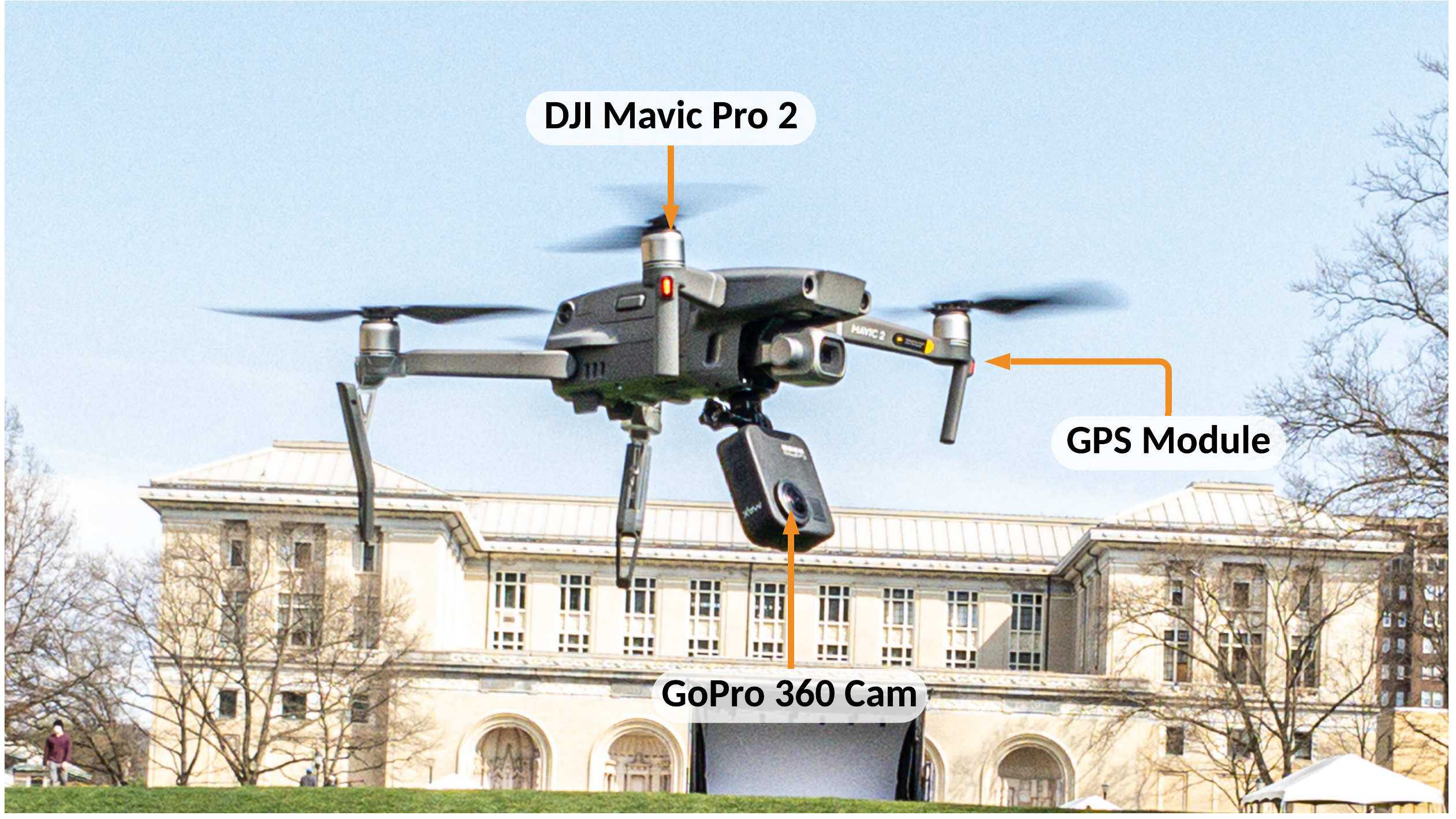}
    	\caption{\textbf{UAV data collection platform.}
    	The platform consists of a DJI Mavic Drone (with GPS), and a omnidirectional camera with an IMU.
        We gather 360$^\circ$ spherical images with time-synchronized IMU and GPS data, under different illumination and viewpoints at Carnegie Mellon University's campus.}
    	\label{fig:platform_UAV}
    \end{figure}

    We evaluate the performance of various VTRN methods with respect to changing illumination, low/high altitudes, and over large flights.
    Two datasets are utilized for comprehensive evaluation: 1) the \textit{CMU Campus} dataset and 2) the \textit{Large Terrain} dataset.
    The \textit{CMU Campus} dataset is designed to test invariance of VTRN methods to different lighting and altitudes, as well as test 3D localization during the take-off/landing phase of a flight.
    This dataset was recorded with a DJI Mavic 2 Pro quadcopter equipped with a GoPro Max omnidirectional camera as shown in Fig.~\ref{fig:platform_UAV}.
    
\begin{figure}[t]
    	\centering
        \includegraphics[width=\linewidth]{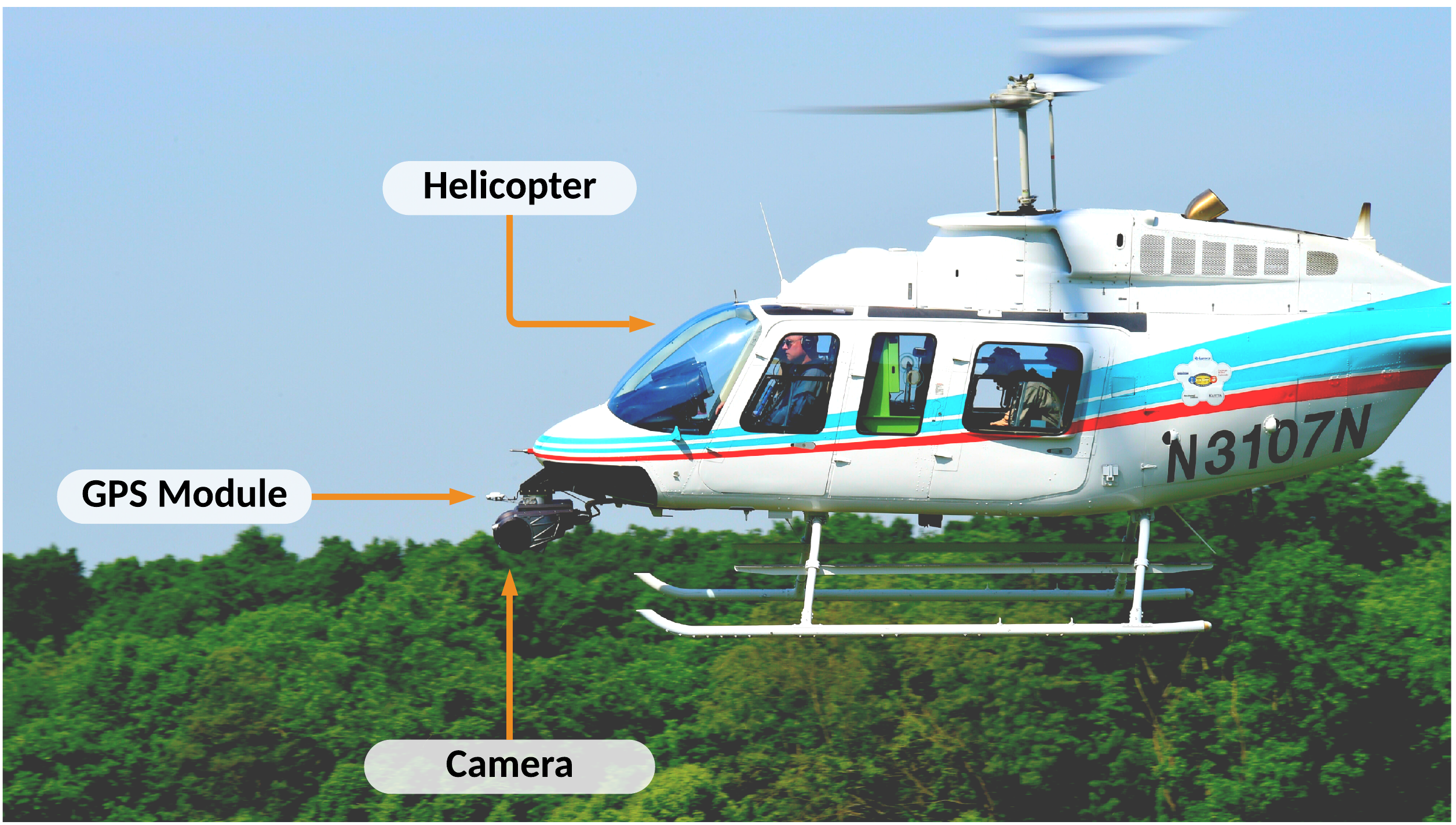}
    	\caption{\textbf{Helicopter platform.}
    	A helicopter equipped with a GPS and a downward-facing pinhole camera is used to collected the $150km$ long trajectory for localization.
    	}
    	\label{fig:platform_Heli}
    \end{figure}

    The \textit{Large-Terrain} dataset contains a $150$km trajectory from Ohio, to Pennsylvania. The focus of this dataset is to capture complex natural terrain over longer trajectories at higher altitudes.  
    This dataset was collected using a helicopter with a downward-facing pinhole camera and a GPS; Fig.~\ref{fig:platform_Heli} shows this platform\footnote{https://allhands.navy.mil/Media/Gallery/igphoto/2002418498/}.
    Using recorded GPS points from the quadcopter and helicopter, we generate corresponding rendered or cropped images from high resolution overhead imagery from Google Earth Engine~\cite{google_engine}.
    
    The following sections will expand on details of our two datasets, evaluation metrics, and comparison methods, and then analyze localization performance of iSimLoc with respect to other methods for three factors: \textit{Condition-invariance}, \textit{Viewpoint-invariance}, and \textit{Global re-localization Performance}.
    
\begin{figure*}[t]
        \centering
        \includegraphics[width=0.9\linewidth]{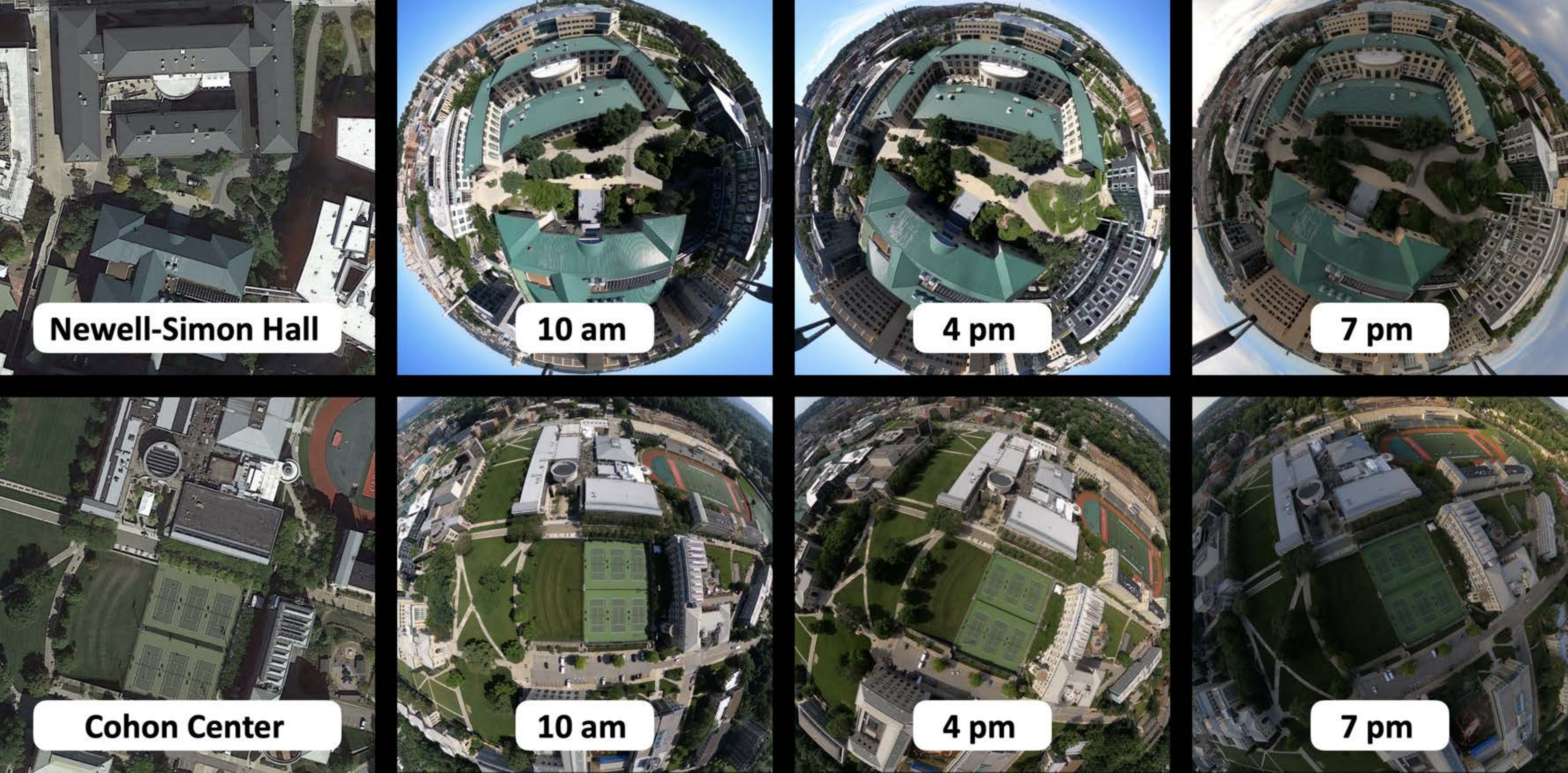}
        \caption{\textbf{Examples of viewpoint and illumination changes in the \textit{CMU Campus} dataset.}
        The first column shows overhead images of CMU's
        \textit{Newell-Simon Hall} and \textit{Cohon University Center}.
        The next three columns show corresponding UAV images during different times of day, viewpoints, and weather conditions.}
        \label{fig:dataset_cmu}
    \end{figure*}

    \bigskip
    \noindent \textbf{Visual Terrain Relative Navigation Datasets} 
    \begin{itemize}
       \item \textbf{CMU Campus} dataset includes $8$ types of trajectories within the campus of Carnegie Mellon University. 
        For each trajectory, we recorded $4$ passes at different times of day ($10am, 1pm, 4pm, 7pm$), along with different orientations ($[0, 45, 90, 180]^{\circ}$) as shown in Fig.~\ref{fig:dataset_cmu}. 
        The average distance covered over all trajectories is $1.6km$. For each pass we collected time-synchronized spherical video and GPS positions of the quadcopter. We created a 3D simulated environment within Microsoft's AirSim~\cite{exp:airsim}, which outputs paired spherical reference images for these trajectories.  Within the simulator we have a 3D model of relevant portions of CMU campus as well as a high resolution overhead image that is used when the quadcopter reaches a high altitude; we generate the reference trajectories using the GPS data collected from real-world passes.  
        We project images onto spherical-views for low flying mode ($\leq 120m$). 
        We crop the spherical images for high flying mode ($\geq 120m$) to avoid the disturbances from a changing sky.
        We use trajectory $\{1\sim 6\}$ for training, and $\{7,8,9,10\}$ for evaluation.
        Data was collected in the month of June 2021.
        \item \textbf{Large Terrain} contains one $150km$ trajectory of a Bell 206 helicopter flying from Cambridge, Ohio, to Pittsburgh, Pennsylvania in August 2017.
        The dataset includes unstructured natural scenarios (e.g., rivers, forests, and plains), structured city-like environments (e.g., houses, streets, and buildings), and also hybrid-structured rural areas (farms).
        We generate paired helicopter images and overhead images based on the time-synchronized GPS data.
        We split the dataset into sub-trajectories, and use $60\%$ for training and $40\%$ for evaluation.
    \end{itemize}

    \begin{figure*}[!th]
        \centering
        \includegraphics[width=0.9\linewidth]{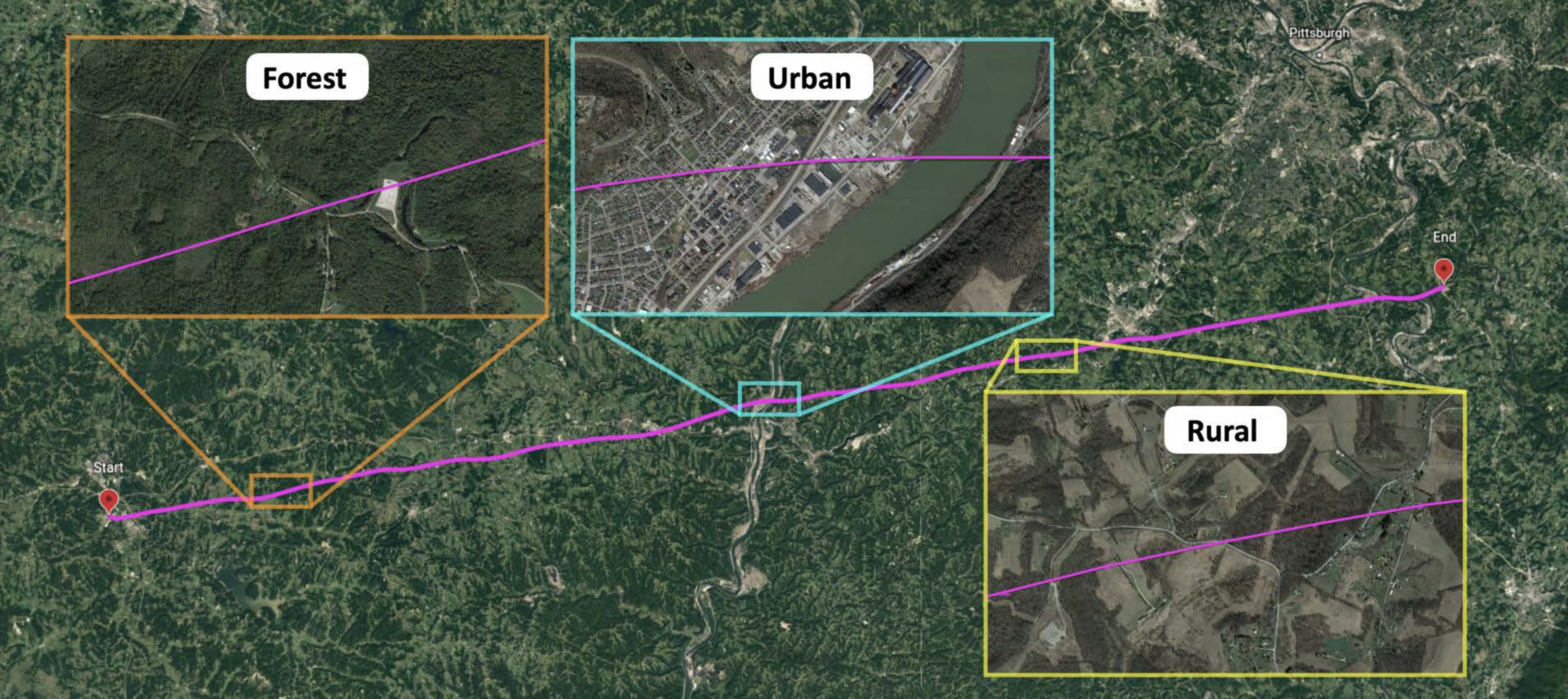}
        \caption{\textbf{The $150$km trajectory flight captured in the \textit{Large Terrain} dataset.}
        The flight covers challenging areas, including natural, urban, and rural terrain.}
        \label{fig:dataset_nea}
    \end{figure*}
    
    \bigskip
    Table.~\ref {table:dataset} shows the differences between our two VTRN datasets.
    The average distance of the \textit{CMU Campus} trajectories is $1,600m$, and each trajectory includes $4$ passes under different time of the day.
    The \textit{Large Terrain} dataset includes $150$ sub-trajectories with average distance of $1,000m$.
    For both datasets, there is no overlap between training and evaluation datasets.
    Fig.~\ref{fig:dataset_cmu} shows some example images of different areas of CMU in the \textit{CMU Campus} dataset.
    Fig.~\ref{fig:dataset_nea} shows example scenarios encountered in the \textit{Large Terrain} flight.

    \begin{table}[t]
        \centering
        \caption{Properties of \textit{CMU Campus} and \textit{Large Terrain} datasets.}
        \begin{tabular}{c c c c}
        \hline
                & \textit{Light Cond.} & \textit{Distance} & \textit{Area} \\ \hline
        \textit{CMU Campus}     & Varying              & $1,600m\times 10 \times 4$           & Urban\\ 
        \textit{Large Terrain}       & Constant             & $1,000m\times 150$            & Urban, Terrain\\ 
        \hline
        \label{table:dataset}
        \end{tabular}
    \end{table}

    \begin{figure*}[ht]
        \centering
        \includegraphics[width=0.98\linewidth]{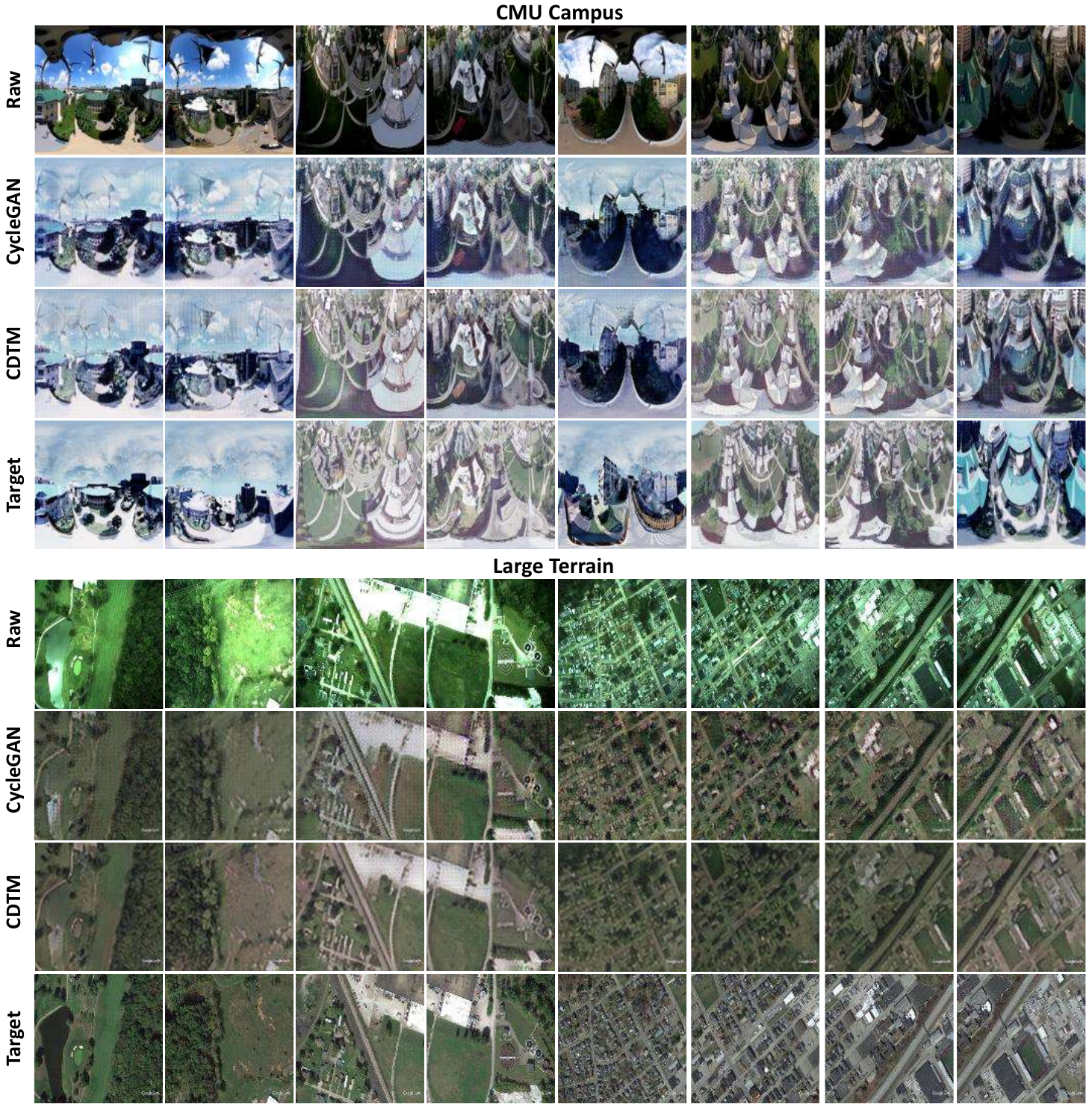}
        \caption{
        \textbf{Condition-invariant Domain Transfer.}
        The two sets of four rows show domain transfer results on \textit{CMU Campus} and \textit{Large Terrain} datasets respectively.
        For each dataset, we demonstrate reconstruction results from CycleGAN and our CDTM module separately.}
        \label{fig:domain_transfer}
    \end{figure*}

    \color{black}
    \bigskip
    \noindent \textbf{Evaluation Methods and Metrics}
    We evaluate the localization performance of our method on both \textit{CMU Campus} and \textit{Large Terrain} datasets via traditional place retrieval metrics.
    The evaluation dataset consists of cross-domain reference and testing queries on the same trajectories. 
    Each testing frame can find its correspondences on reference queries. 
    Successful place feature matching is based on testing queries' retrieval poses.
    If the deviation distance of retrieval and target query is within a given threshold ($20$m for \textit{CMU Campus} dataset and $40$m for \textit{Large Terrain} dataset), place recognition is counted as successful; otherwise, it is unsuccessful.
    We use average recall of top $10$ and top-N retrievals, receiver operating characteristic curve, feature difference, relative orientation distributions, and global re-localization success rates to analyze place recognition accuracy on evaluation trajectories of both datasets.
    We compare iSimLoc with learning-based feature learning methods (CycleGAN~\cite{CycleGAN2017}, AlexNet~\cite{loc:Alexnet}, NetVLAD~\cite{PR:netvlad}, CALC~\cite{VPR:CALC}), and non-learning based geometric methods HoG~\cite{VPR:HoG}, Bag-of-words (BoW)~\cite{FeatureCapturer:BoW2} and CoHoG~\cite{vpr:co_hog}).
    
    All learning-based methods are trained with the same hardware: an Ubuntu 18.04 system with $64$GB of RAM and one Nvidia 1080Ti GPU. 
    Default visual input dimensions for all methods is set to $256\times256$.
    To train iSimLoc, we use only $10\%$ of paired sim/real data from each dataset for domain-transfer training, and we use $60\%$ of the overhead images for place recognition training; we evaluate on cross sim/real domains with the remaining $40\%$ of the data.
    For other methods, we use $60\%$ for training, and leave $40\%$ as unseen environments for evaluation.

    In condition-invariant analysis, we fix viewpoints and calculate place recognition average recalls of different methods under changing conditions.
    In viewpoint-invariant analysis, we fix environmental conditions, and calculate average recall under different viewpoints (translations, orientations, and altitude ratios).
    In hierarchical localization analysis, we analyzed successful global re-localization rates on both datasets using different global localization methods.
    
    \begin{figure*}[!th]
        \centering
        \includegraphics[width=\linewidth]{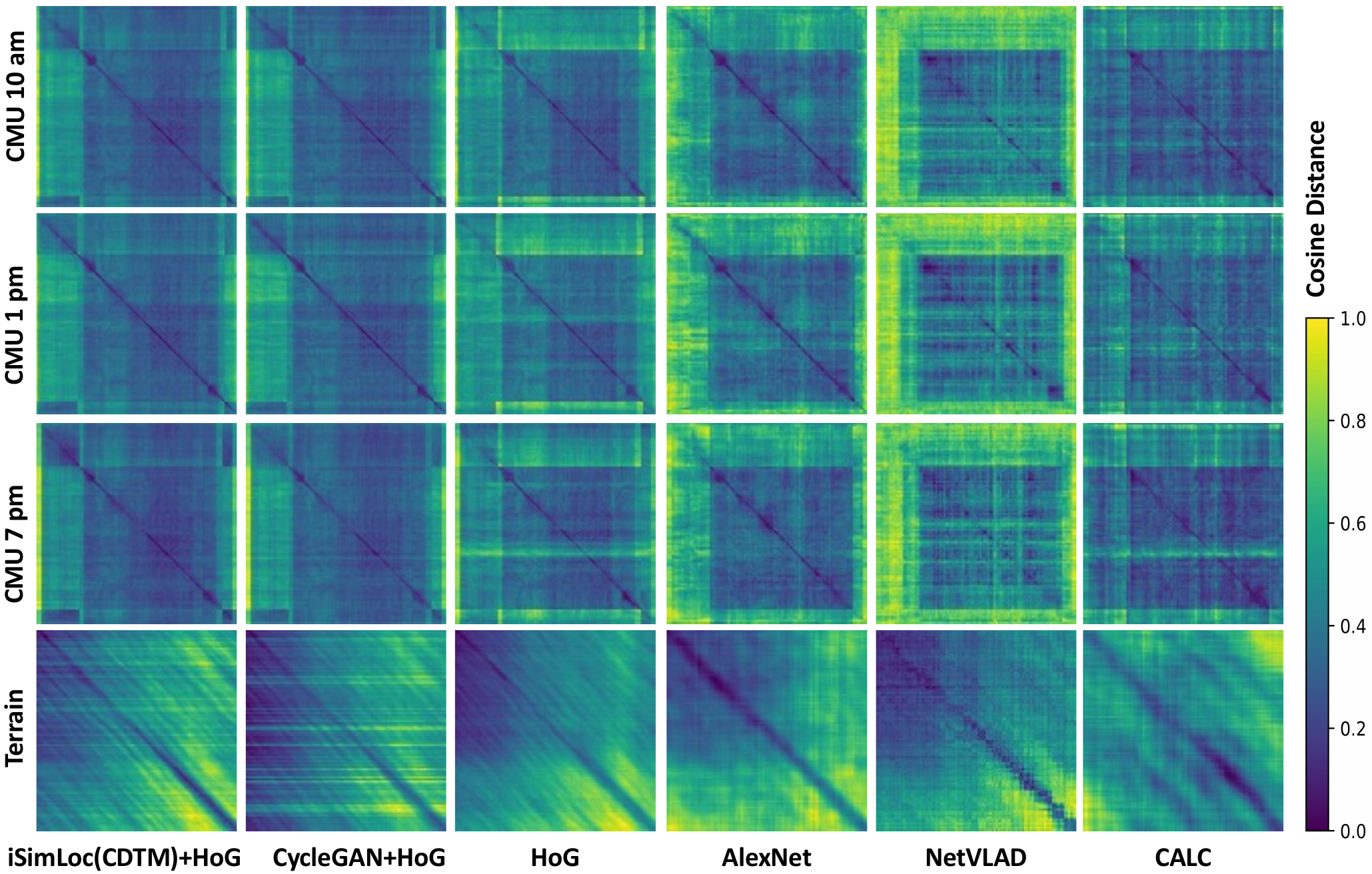}
        \caption{\textbf{Real-to-Sim feature differences for different datasets.}
        Each sub-figure represents the difference matrix between real images (x-axis) and corresponding overhead images (y-axis). Similarities are calculated by cosine distance.
        The first three rows show results on the same trajectory of \textit{CMU Campus} datasets under illuminations of $[10am, 1pm, 7pm]$.
        The last row shows matching results on \textit{Large Terrain} dataset.
        }
        \label{fig:feature_diff}
    \end{figure*}
    
    \begin{figure*}[!th]
        \centering
        \includegraphics[width=\linewidth]{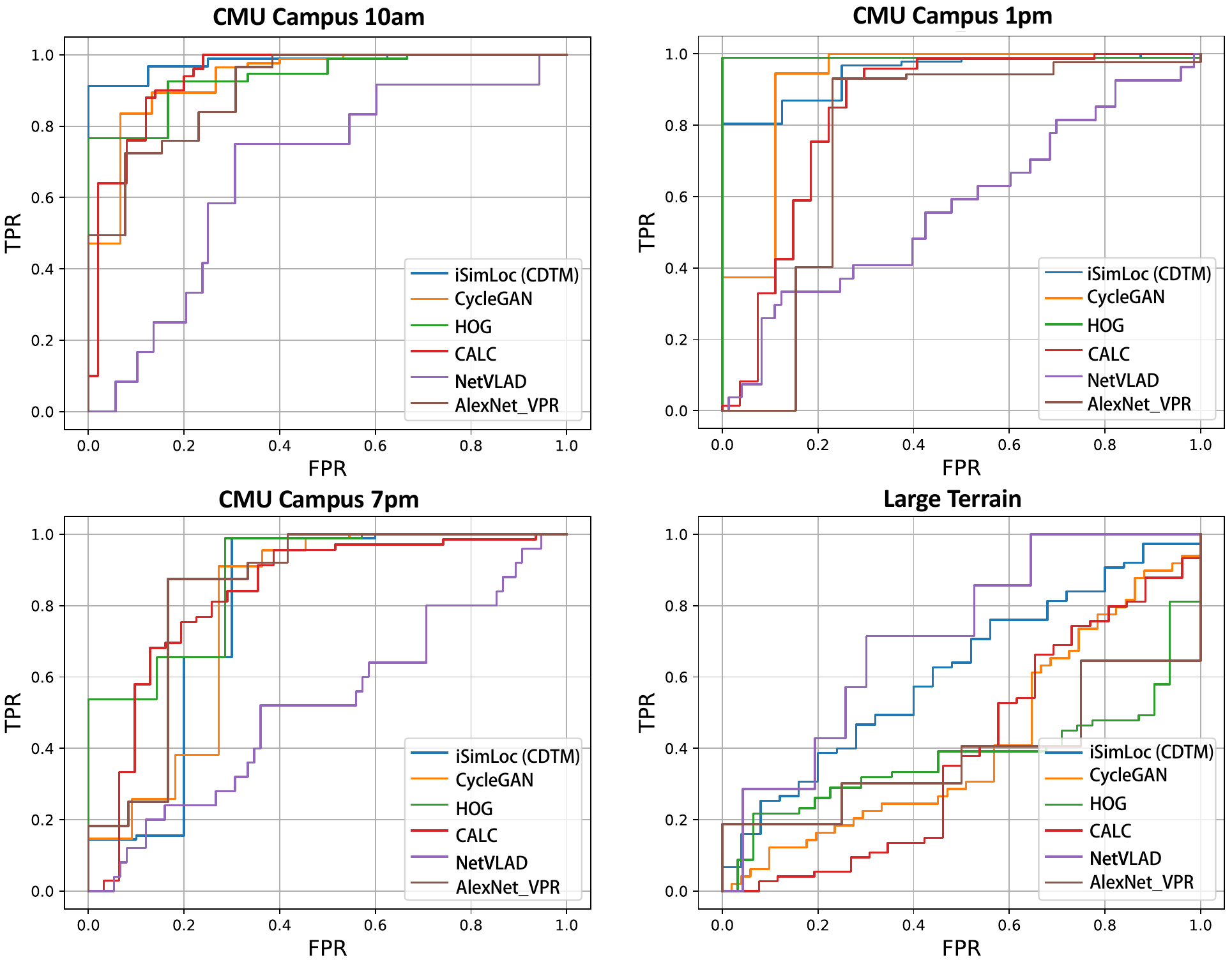}
        \caption{\textbf{ROC analysis for cross-domain place recognition.}
        Each sub-figure shows the receiver operating characteristic (ROC) curves of different methods for unseen environments on \textit{CMU Campus} under conditions $[10am, 1pm, 7pm]$ and \textit{Large Terrain} datasets.}
        \label{fig:ROC_curve}
    \end{figure*}    

    \subsection{Domain-invariant Place Recognition Analysis}
    \label{sec:condition_analysis}
    This section evaluates visual place recognition performance for different environmental conditions.
    We provide raw images and paired overhead images for the same perspectives for a fair comparison.
    Furthermore, we compare our \textit{CDTM} module with existing non-learning method HoG to verify place recognition ability.
    We first show domain adaptation on both \textit{CMU Campus} and \textit{Large Terrain} datasets by transferring raw images into overhead images.
    Based on \textit{CDTM} module's orthogonal feature constraints, we extract paired geometry features ($Z_G$) and condition features ($Z_C$) from the same image.
    As stated in Section.~\ref{sec:CDTM}, additional conditional features from overhead images assists in domain transfer of real images.
    We compared recognition ability with and without orthogonal extraction to verify the above property.
    
    Fig.~\ref{fig:domain_transfer} shows domain transfer on unseen environments.
    After training the network model on our two datasets, we pick unseen trajectories to examine image reconstruction performance.
    The first rows of both sections in the figure show the input images; the last rows show the target images, and the third rows show images generated with our \textit{CDTM} module.
    For comparison, we also show images generated with CycleGAN~\cite{CycleGAN2017} network modules in the second rows.
    Different from CycleGAN, under varying illumination of \textit{CMU Campus} dataset, our \textit{CDTM} module also encodes condition styles into the image reconstruction.
    CycleGAN requires that target images must follow the same style, which restricts image reconstructions on unseen environments or conditions.
    Fig.~\ref{fig:feature_diff} shows feature difference matrices between test query images and reference overhead images.
    We notice that both non-learning and learning methods have varying performance if conditions change. In general, AlexNet and methods with domain transfer modules show higher robustness.
    With different domains as constraints, NetVLAD shows less generalization ability than AlexNet. However, the large size of the AlexNet model makes it ill-suited for real-time inference.
    With a domain transfer module (\textit{CDTM} and CycleGAN), HoG shows higher robustness to domain differences.
    When comparing two different domain transfer modules, our \textit{CDTM} module is able to generalize better than CycleGAN for unseen environments.
    Amongst learning-based methods, CALC and original NetVLAD are more sensitive to lighting changes. They cannot capture rich geometric structures for proper place retrievals, especially on the \textit{Large Terrain} dataset.

    \begin{table}[t]
        \centering
        \caption{\textbf{The average recall of top $@10$ retrievals for different environments.}}
        \begin{tabular}{ |c | c | c | c |c |}
        \hline
        Method & \textit{C10am} & \textit{C1pm} & \textit{C7pm} & \textit{Terrain} \\ \hline
        NetVLAD~\cite{PR:netvlad}    & $46.8\%$ & $52.1\%$  & $36.4\%$ & $68.1\%$ \\ \hline
        CALC~\cite{vpr:co_hog}       & $64.5\%$ & $70.8\%$  & $50.2\%$ & $80.2\%$ \\ \hline
        AlexNet~\cite{loc:Alexnet}   & \textHT[98.5\%] & \textLT[95.8\%]  & $76.0\%$ & \textLT[90.7\%] \\ \hline 
        HoG~\cite{VPR:HoG}           & $80.2\%$ & $87.5\%$  & $75.0\%$ & $86.5\%$ \\ \hline
        CycleGAN~\cite{CycleGAN2017}+HoG & $93.7\%$ & $94.9\%$  & \textLT[88.5\%] & $89.3\%$ \\ \hline
        iSimLoc(CDTM)+HoG                & \textLT[96.5\%] & \textHT[98.3\%]  & \textHT[93.8\%] & \textHT[92.8\%] \\ \hline
        \end{tabular}
        \label{table:all}
    \end{table}
    
    Table.~\ref{table:all} shows analysis of average recall for top $10$ retrievals for different methods; domain transfer based on CycleGAN and CDTM  further improves place retrieval accuracy compared to other methods.
    We also notice that AlexNet performs well at the $10am$ case from the \textit{CMU Campus} dataset, while CDTM performs well in all conditions.
    In Fig.~\ref{fig:ROC_curve}, we present the receiver operating characteristic (ROC) curves of different methods.
    CDTM has robust true-positive rates compared with other methods.
    However, while the non-learning HoG method also performs well on $1pm$ \textit{CMU Campus} dataset, it is not robust for all time conditions.
    Overall, the CDTM-based domain transfer module provides a high average recall rate with top $10$ retrievals, and has a robust and confident estimation of potential matches.

    \begin{figure*}[t]
        \centering
        \includegraphics[width=0.918\linewidth]{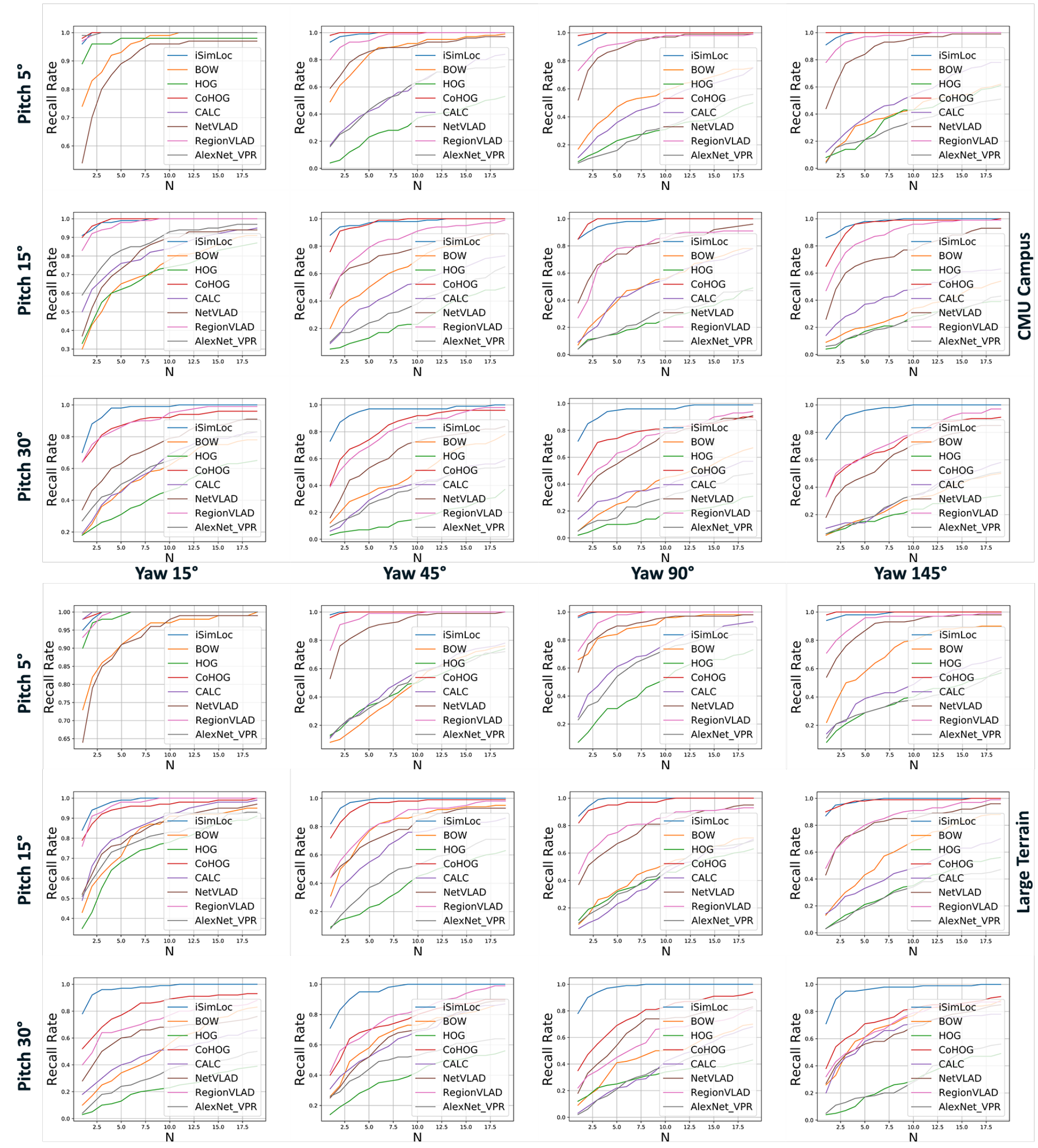}
        \caption{\textbf{Localization results for different viewpoints on different datasets.}
        For each dataset, we pick one trajectory from the same domain and generate test/reference queries with different pitch angles $[5,15,30]^{\circ}$ and yaw angles $[15,45,90, 135]^{\circ}$, and then analyze the average recall for top-$N$ retrievals.
        }
        \label{fig:viewpoint_invariant}
    \end{figure*}

   \begin{figure*}[t]
        \centering
        \includegraphics[width=0.49\linewidth]{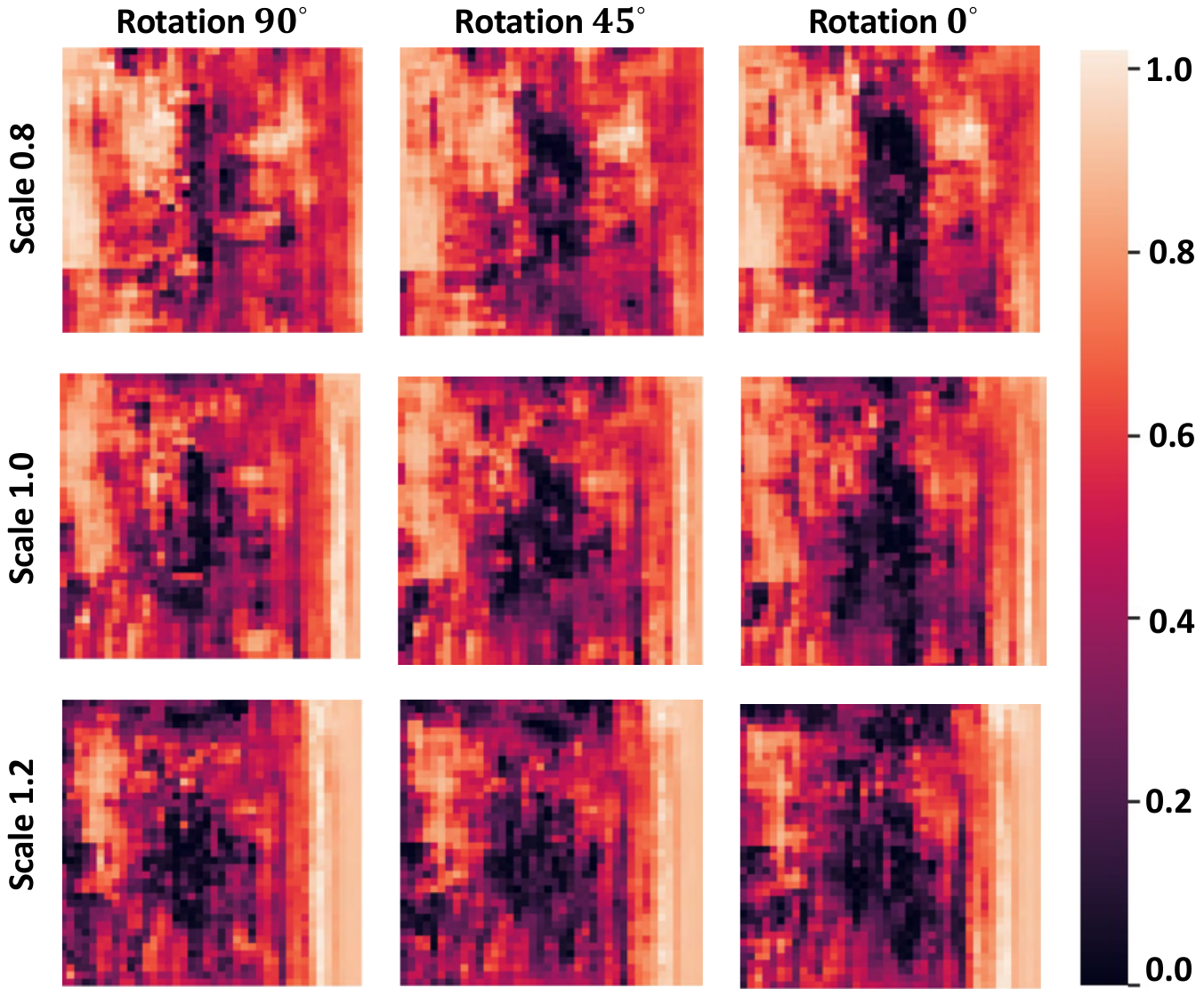}
        \includegraphics[width=0.49\linewidth]{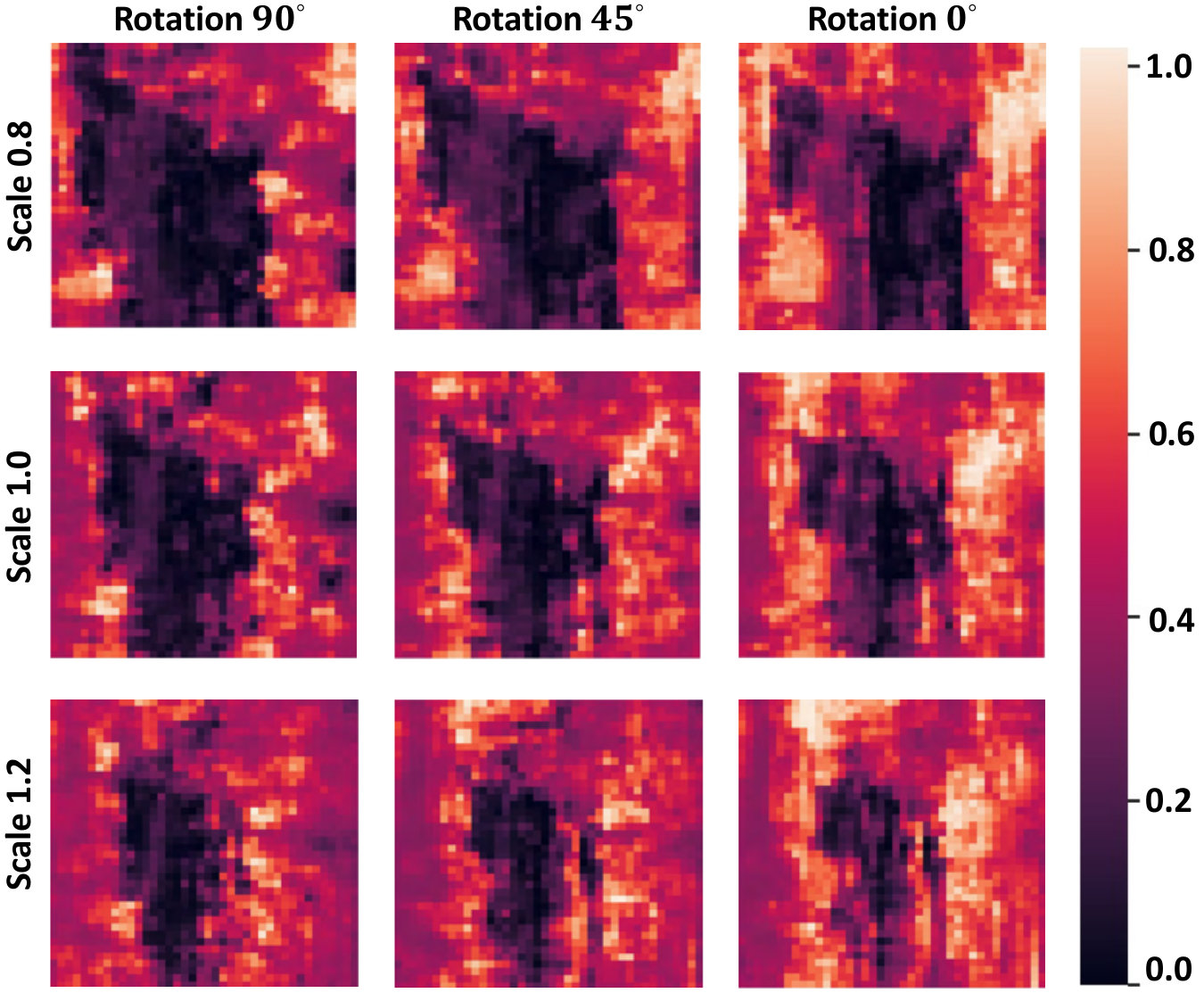}
        \caption{\textbf{Localization results at high altitude with various viewpoints and scales.}
        In each sub-figure, we compare feature distance between fixed visual images and their corresponding simulated images.
        $x$ and $y$ axes represent offset distance on the XY-plane $[64\times 64]$. Each pixel is equal to $1m$.
        }
        \label{fig:exp_viewpoint}
    \end{figure*}

    \begin{figure}[t]
        \centering
        \includegraphics[width=\linewidth]{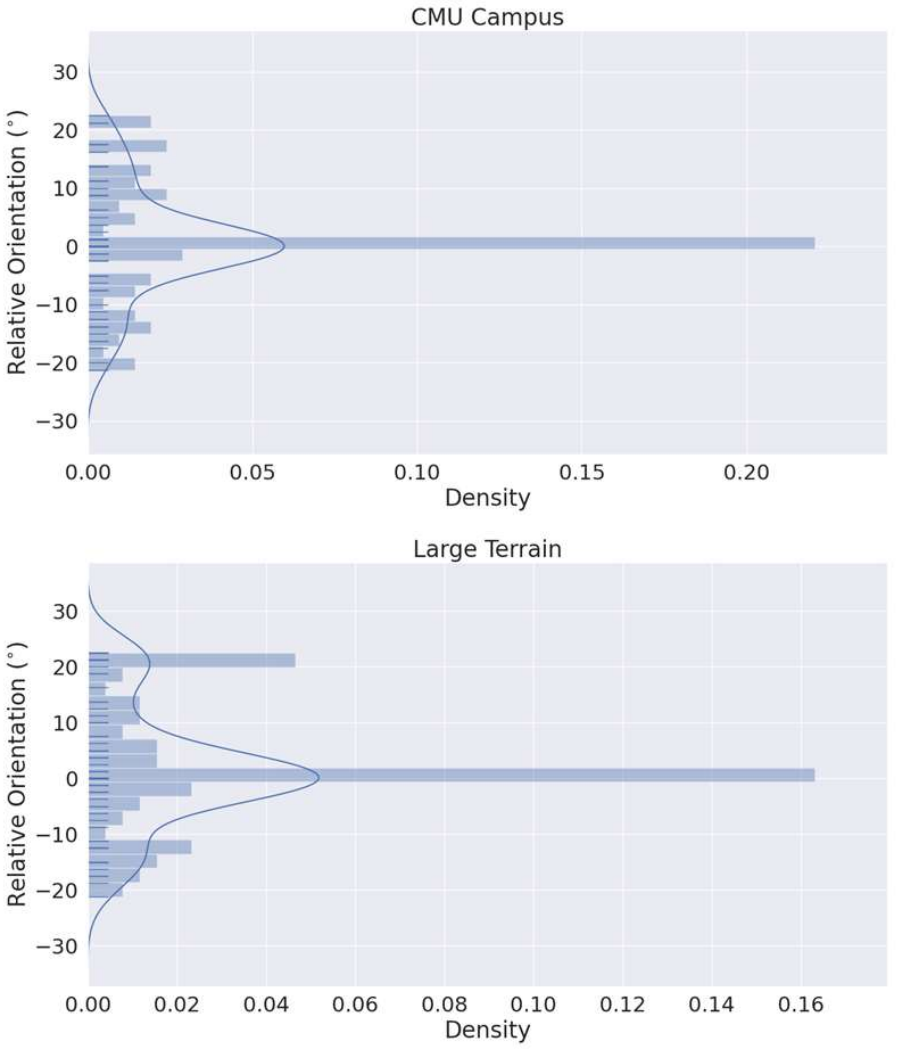}
        \caption{\textbf{Orientation estimation on our two datasets.}
        For each dataset, we manually set the orientation difference between testing and reference queries with a default yaw offset of $\mathcal{N}\sim(0,90^\circ)$, and analyze the relative orientation estimation of the PSE module.}
        \label{fig:exp_orientation}
    \end{figure}

    \subsection{Viewpoint-Invariant Place Recognition Analysis}
    \label{sec:VI}

    On both \textit{CMU Campus} and \textit{Large Terrain} datasets, our pose estimation module (\textit{PEM}) extracts viewpoint-invariant descriptors for place retrievals from extracted condition-invariant features from our (\textit{CDTM}) module.
    For this experiment, both reference and test images are from same domain to focus on robustness to viewpoint differences.
    Our place descriptor constrains global localization to the Euclidean domain.
    Moreover, given matching testing and reference images, the \textit{PEM} estimates relative orientations in parallel.
    In this subsection, we investigate robustness to viewpoint differences and accuracy of orientation estimation separately.

    Firstly, as shown in Fig.~\ref{fig:viewpoint_invariant}, for each dataset we analyze viewpoint-invariance by applying several orientations on yaw ($[15^{\circ}, 45^{\circ}, 90^{\circ}, 145^{\circ}]$) and pitch ($[5^{\circ}, 15^{\circ}, 30^{\circ}]$) angles to testing images.
    Most place recognition methods show high retrieval rates but only for small viewpoint changes.
    For a fixed pitch angle, place recalls of different methods drops significantly as the yaw angle is changed.
    As the yaw angle changes, iSimLoc shows higher and more consistent performance.
    With only viewpoint differences, iSimLoc's top $5$ place retrievals on both datasets is above $90\%$.

    We further analyze viewpoint-invariance from a different perspective.
    For both datasets, we analyze the similarity of iSimLoc features given reference images from a local area with different rotations and scales.
    For each sub-figure of Fig.~\ref{fig:exp_viewpoint}, an image pixel corresponds to relative translation difference, and the center of the images are ground truth matching areas.
    We take reference images for different field of views, e.g. scale $1.0$ means reference and test images have the same perspective.
    Scale $0.8$ and $1.2$ indicate reference images are taken at $80\%$ and $120\%$ of the altitude of testing images.
    For local translation, orientation, and scale differences, iSimLoc shows a higher similarity score on both \textit{CMU Campus} and \textit{Large Terrain} datasets, which will improve global re-localization robustness for scale differences.

    Additionally, given a matched image pair, relative orientation estimation uses the same spherical features that are used for the viewpoint-invariant descriptor as depicted in Section.~\ref{sec:OE}.
    As shown in Fig.~\ref{fig:exp_orientation}, on both \textit{CMU Campus} and \textit{Large Terrain} datasets, relative orientation between testing and reference queries can be evaluated according to maximum spherical correlation $\hat{C}$ as we stated in section~\ref{sec:OE}.
    For both datasets, we present the relative orientation error distribution, which has a domain within $[-30\sim 30]^{\circ}$.
    Compared to performance on \textit{CMU Campus} dataset, the estimator shows higher orientation error on the \textit{Large Terrain} dataset, and this is mainly caused by textureless terrain environments, which increase difficulty of accurate orientation estimation.
    
    \begin{figure*}[!th]
        \centering
        \includegraphics[width=\linewidth]{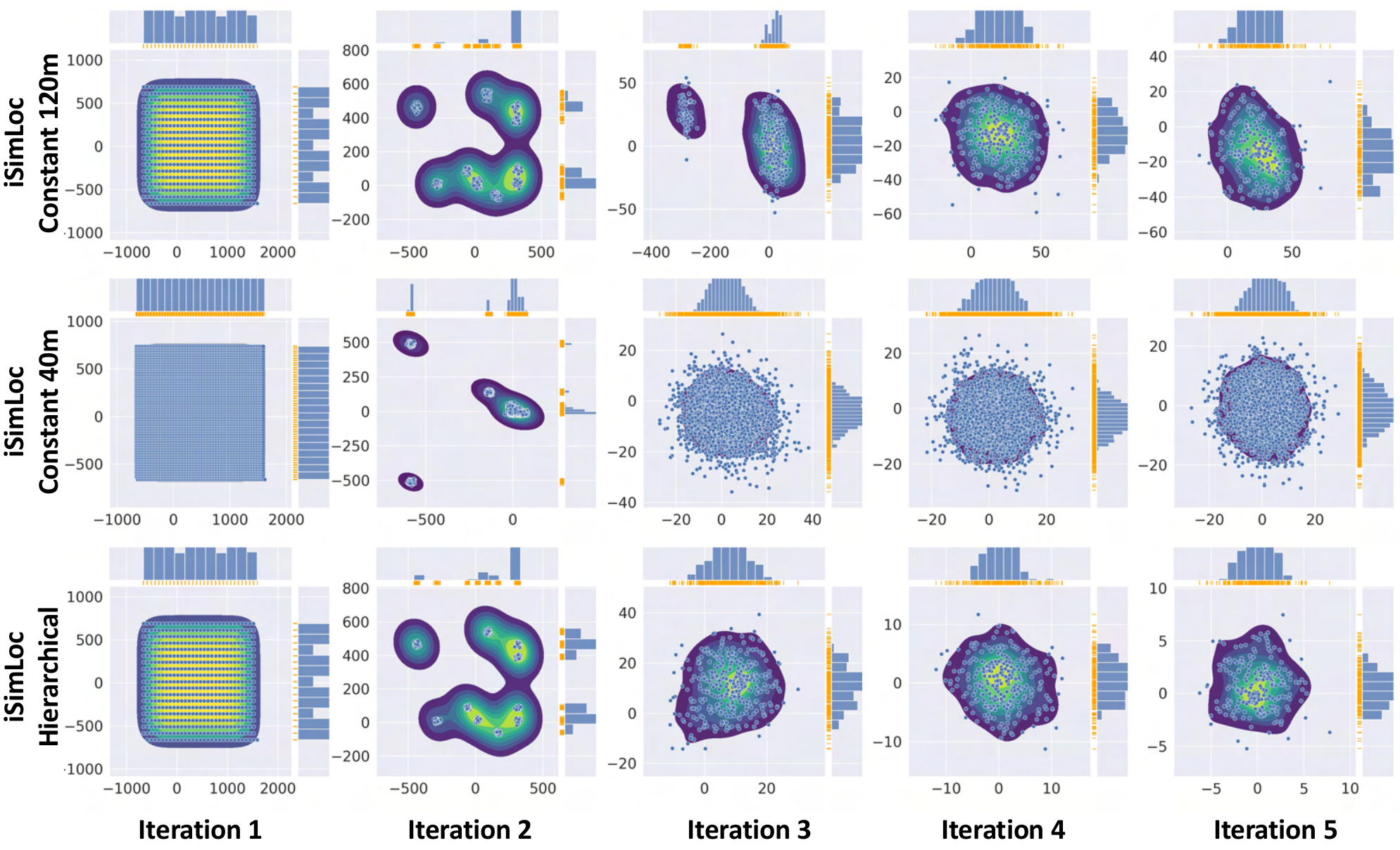}
        \caption{\textbf{Comparison of different global localization methods on \textit{CMU dataset}.}
        The first two rows show global localization with iSimLoc place features at constant altitudes ($120m$ and $40m$ respectively).
        The last row shows localization using a  hierarchical manner (from a $120m$ perspective).
        In each sub-figure, we plot particle distributions and density analysis. The X-Y positions are relative to the target position.
        }
        \label{fig:CMU_hiloc}
    \end{figure*}

    In general, the \textit{PEM} module provides accurate viewpoint-invariant place recognition, which improves global localization efficiency by reducing the sampling space from 6-DoF space to Euclidean space.
    The orientation estimation module concurrently estimates relative orientation between testing images and matched query images. 
    However, as we can see in the feature matching results, there are outliers due to common and similar textures, which is something that becomes more abundant in long-term navigation. 
    In the following subsection, we investigate hierarchical localization performance in terms of robustness, accuracy, and efficiency.

    \subsection{Hierarchical Localization Analysis}
    \label{sec:OLP}
    Hierarchical matching improves global localization efficiency over a large search area without losing matching accuracy, as depicted in Section.~\ref{sec:hiloc}.
    Robustness of hierarchical localization can be further boosted by ignoring condition and viewpoint changes, which is done with the aid of the domain transfer module (\textit{CDTM}) and viewpoint-invariant module (\textit{PEM}).
    We analyze different global localization methods on the \textit{CMU Campus} dataset as shown in Fig.~\ref{fig:CMU_hiloc}, i.e., fixed $120m$ and $40m$ altitudes, and high-to-low altitudes.
    Global localization is done through using  the same iSimLoc place feature and particle filter, under the same iteration times and resampling mechanism.
    For each iteration, we only keep $80\%$ of the original particles.
    Matching using a fixed altitude is often sub-optimal since one has to trade between accuracy and amount of context captured within place features.
    In contrast, hierarchical localization starts from a broad field of view (FOV) to find potential best matches, and incrementally changes perspectives by cropping higher resolution regions of the image.
    This matching method helps iSimLoc balance search efficiency and robustness, which is especially important for large-scale re-localization tasks.
    
    We also evaluate matching accuracy and efficiency by investigating recognition rates and matching time on our two datasets.
    For the \textit{CMU Campus} dataset, we take the whole campus area as the reference map, which covers an area of $700m \times 400m$.
    The testing image is taken from an altitude of $120m$ with an omnidirectional camera.
    For the \textit{Large Terrain} dataset, we take a selected area as reference map, which covers a space of $1000m \times 500m$.
    The test image is taken from an altitude of $200m$ with a pin-hole camera pointing towards the ground.
    A successful recognition is defined as having a matching distance within $20m$ for \textit{CMU Campus} and $40m$ for \textit{Large Terrain}.
    To maintain the matching efficiency, all the reference image features are pre-stored.
    Table.~\ref{table:hiloc} shows matching performance for different matching modes, when given different numbers of initial particles.
    When using constant crops for global localization, iSimLoc cannot provide high recognition rates with fewer initial particles. However, matching with more initial particles requires more processing time.
    With hierarchical matching we are able to match well even with few initial particles.
    Specifically, at a higher initial altitude, the hierarchical matching approach enables a more robust initial estimate, which  further helps achieve successful localization.
    Regarding the efficiency, we notice that a hierarchical approach speeds up localization by a factor of $4\sim 10$ times.
    Compared with the performance on \textit{CMU Campus} dataset, localization over the \textit{Large Terrain} dataset shows lower success rates, which is likely caused by less-distinguishable geometric features in the terrain.

       \begin{table*}[t]
        \centering
        \caption{\textbf{Global re-localization robustness and efficiency.}}
        \begin{tabular}{ |c |c|c|c|c| c | c |c | c| c | }
        \hline
         & \textit{$ACC_{0.9}$} & \textit{$ACC_{0.8}$} & \textit{$ACC_{0.7}$} & \textit{$ACC_{0.9}$} & \textit{$ACC_{0.8}$}  & \textit{$ACC_{0.7}$} & \textit{$Time_{0.9}$} & \textit{$Time_{0.8}$} & \textit{$Time_{0.7}$} \\ \hline
        {Localization Method} & \multicolumn{3}{c|}{CMU Campus Dataset} & \multicolumn{3}{c|}{Large Terrain Dataset} & \multicolumn{3}{c|}{Run Time} \\ \hline
        & \multicolumn{9}{c|}{Hierarchical Localization} \\ \hline
        CALC~\cite{VPR:CALC} ($100\%$ Altitude) & $10.1\%$ & $6.8\%$  & $4.5\%$ & $10.8\%$ & $7.1\%$ & $4.1\%$  & $2.4s$ & $1.9s$ & $1.6s$ \\ \hline
        HOG~\cite{VPR:HoG} ($100\%$ Altitude) & $12.1\%$ & $9.4\%$  & $5.2\%$ & $11.4\%$ & $8.5\%$ & $5.6\%$  & $2.1s$ & $1.7s$ & $1.5s$ \\ \hline 
        NetVLAD~\cite{PR:netvlad} ($100\%$ Altitude) & $24.7\%$ & $18.3\%$  & $12.4\%$ & $22.5\%$ & $16.8\%$ & $13.6\%$  & $3.1s$ & $2.5s$ & $2.1s$ \\ \hline
        AlexNet~\cite{loc:Alexnet} ($100\%$ Altitude) & $45.8\%$ & $31.6\%$  & $23.5\%$ & $39.7\%$ & $28.4\%$ & $19.2\%$  & $3.3s$ & $2.6s$ & $2.2s$ \\ \hline
        iSimLoc ($100\%$ Altitude) & $88.7\%$ & \textLT[84.5\%]  & $74.8\%$ & $83.8\%$ & $76.1\%$ & \textLT[72.4\%]  & $1.5s$ & $1.2s$ & $0.9s$ \\ \hline 
        iSimLoc ($50\%$ Altitude) & \textLT[89.1\%] & $84.2\%$  & \textHT[78.4\%] & \textLT[85.4\%] & \textLT[79.2\%] & $70.3\%$  & $6.2s$ & $5.0$ & $4.3s$ \\ \hline
        iSimLoc ($33\%$ Altitude) & \textHT[91.8\%] & \textHT[85.3\%]  & \textLT[76.1\%] & \textHT[86.9\%] & \textHT[81.7\%] & \textHT[72.5\%]  & $12.9s$ & $10.2s$ & $9.0s$ \\ \hline
        & \multicolumn{9}{c|}{Constant Altitude Localization} \\ \hline
        iSimLoc ($100\%$ Altitude) & $59.2\%$ & $51.9\%$  & $45.8\%$ & $52.5\%$ & $48.8\%$ & $43.1\%$  & $1.5s$ & $1.2s$ & $1.1s$ \\ \hline
        iSimLoc ($50\%$ Altitude)  & $61.6\%$ & $52.5\%$  & $48.2\%$ & $54.7\%$ & $49.2\%$ & $45.3\%$  & $6.2s$ & $5.1s$ & $4.4s$ \\ \hline
        iSimLoc ($33\%$ Altitude)  & $63.5\%$ & $55.3\%$  & $49.7\%$ & $53.6\%$ & $51.4\%$ & $45.9\%$  & $13.1s$ & $10.8s$ & $9.1s$ \\ \hline
        \end{tabular}
        \label{table:hiloc}
    \end{table*}
    
    \section{Discussion}
    \label{sec:discussion}
    As depicted in System Overview (Section.~\ref{sec:system_oveview}), our iSimLoc method consists of a condition- and viewpoint- invariant place feature learning module, and a hierarchical localization module for \textit{VTRN} task.
    The conditional domain transfer module of iSimLoc improves localization accuracy with the assistance of conditional features of target domains as demonstrated by results shown in Fig.~\ref{fig:domain_transfer}.
    As depicted in Section.\ref{sec:condition_analysis}, our CDTM reconstruction is based on both geometric and conditional features, which further improves localization accuracy for unseen environments compared to CycleGAN and AlexNet by $3.75\%$ and $5.09\%$ on average.
    However, the current domain transfer module can only be used with low-resolution images ($128\times 128$ or $256\times 256$), which reduces it's ability to capture rich geometric features.
    The Pose Estimation Module (PEM) that is shown in Section.\ref{sec:VI} demonstrates robustness to viewpoint differences and orientation estimation compared to other methods by $5\sim10\%$ as depicted in Fig.~\ref{fig:viewpoint_invariant}.
    However, the current spherical convolution network in the \textit{PEM} has very shallow layers ($4$ spherical convolutions), which may reduce its feature extraction ability.
    Finally, as analyzed in Section.\ref{sec:hiloc}, our hierarchical searching method balances matching efficiency and accuracy.
    As shown in Table.~\ref{table:hiloc}, by using the hierarchical method, iSimLoc can reach up to $80\%$ successful retrieval rates for large-scale place recognition compared to $40\%$ for the next best method AlexNet. 
    
    Since iSimLoc is designed to give the top matches for global re-localization, for a complete system it needs to be further combined with online image alignment methods for accurate pose estimation, and visual odometry for continuous pose estimation.
    As shown in the run time analysis in  Table.~\ref{table:hiloc}, the hierarchical version of iSimLoc has $88.7\%$ (Campus) and $83.8\%$ (Large Terrain) correct matches with a $1.5s$ compution time, and $89.1\%$ and $85.4\%$ correct matches  with a $6.2s$ computation time.


    \section{Conclusion}
    \label{sec:conclusion}
    This paper presented iSimLoc, a hierarchical global re-localization method for visual terrain relative navigation (VTRN) with the assistance of overhead imagery. iSimLoc can learn a place recognition model with only simulated images and only requires a small portion ($10\%$) of paired sim/real data to train the domain-transfer module compared to other methods that require $60\%$ of the data for domain transfer.
    Since it is viewpoint-invariant property it recognizes the same place even with orientation and altitude differences, and
    the hierarchical matching method helps iSimLoc balance efficiency and robustness for global re-localization.

    In future research, we intend to improve the efficiency of hierarchical localization by parallelizing the method.
    Additionally, we also plan to improve the current domain-transfer module's ability to capture rich geometric details so that we can integrate iSimLoc with existing image alignment methods for accurate pose estimation.
    
\section*{Acknowledgments}
The authors gratefully acknowledge Near Earth Autonomy, Pittsburgh, PA, for assistance in creating the ``\textit{Large Terrain}'' dataset.

\bibliographystyle{IEEEtran}
\bibliography{bible}

\vspace{-1cm}
\begin{IEEEbiography}[{\includegraphics[width=1in,height=1.25in,clip,keepaspectratio]{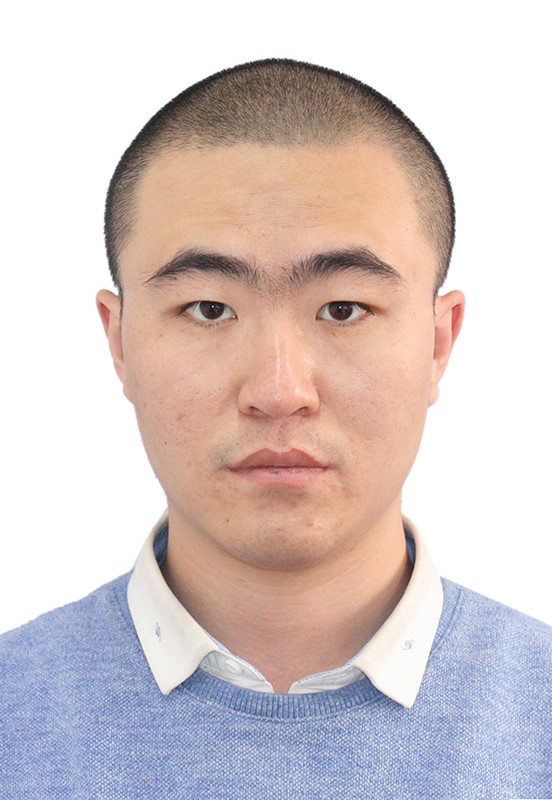}}]
    {Peng Yin} received his Bachelor degree from Harbin Institute of Technology, Harbin, China, in 2013, and his Ph.D. degree from the University of Chinese Academy of Sciences, Beijing, in 2018.
    He is a research Post-doctoral with the Department of the Robotics Institute, Carnegie Mellon University, Pittsburgh, USA.
    His research interests include LiDAR SLAM, Place Recognition, 3D Perception, and Reinforcement Learning. Dr. Yin has served as a Reviewer for several IEEE Conferences ICRA, IROS, ACC, RSS.
\end{IEEEbiography}

\begin{IEEEbiography}[{\includegraphics[width=1in,height=1.25in,clip,keepaspectratio]{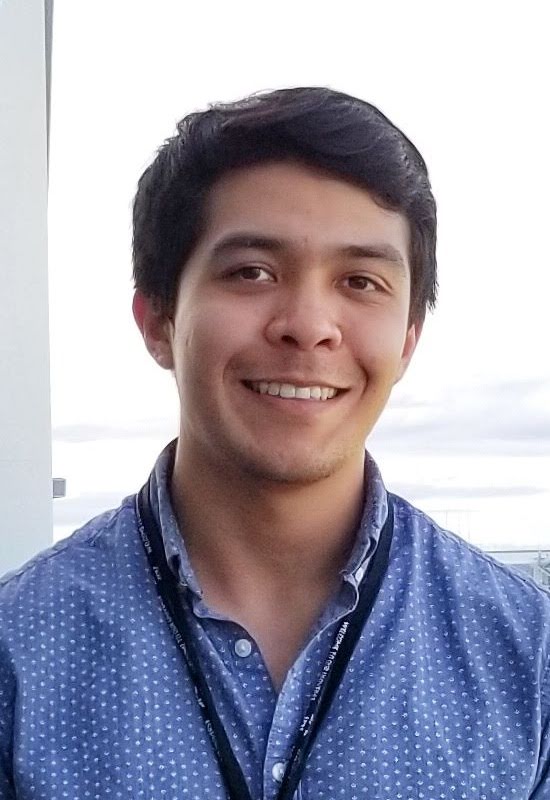}}]
    {Ivan Cisneros} received his B.S. in Electrical Engineering with a minor in Computer Science from Harvard University in 2016.
    He worked full time at NASA-JPL on several flight projects for 3 years before starting his graduate studies at Carnegie Mellon University. He is currently working on a Master's degree in Robotics within the Robotics Institute at CMU.
    His research interests include SLAM, visual localization, 3D Perception, and Deep Learning.
\end{IEEEbiography}

\begin{IEEEbiography}[{\includegraphics[width=1in,height=1.25in,clip,keepaspectratio]{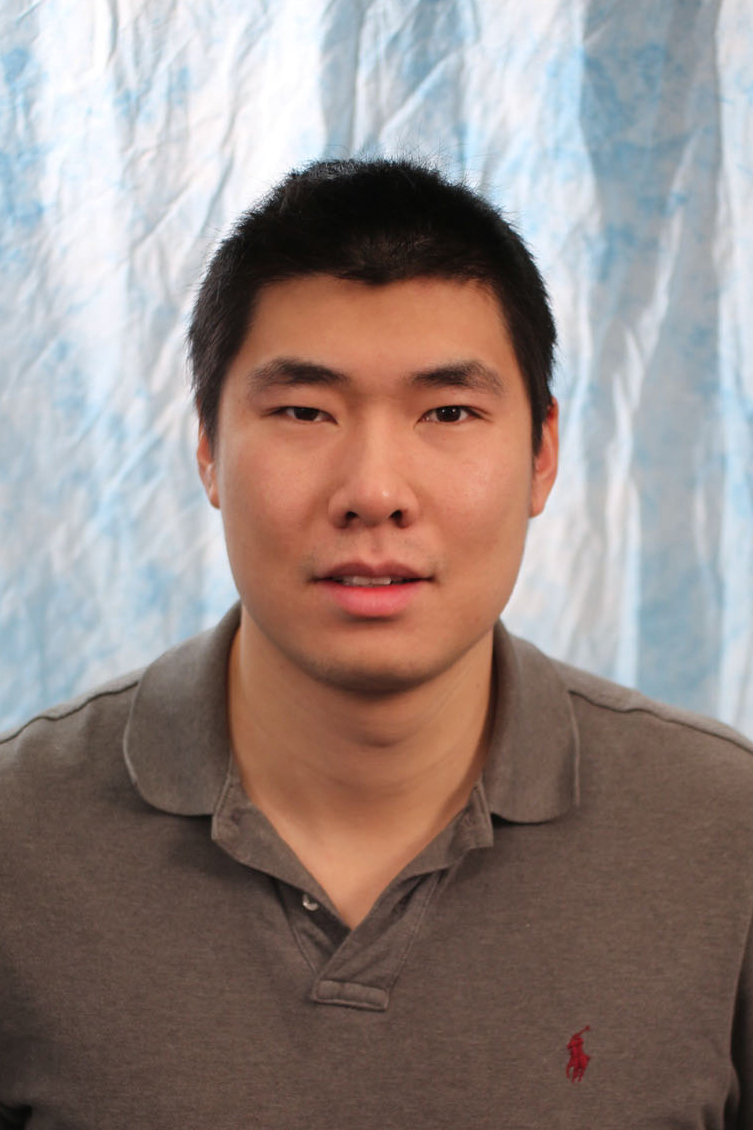}}]
    {Ji Zhang} received his Ph.D. in Robotics from Carnegie Mellon University in 2017.
    Ji Zhang is a Systems Scientist at the Robotics Institute at Carnegie Mellon University, where he leads in the development of a series of autonomous navigation algorithms. His work was ranked \#1 on the odometry leaderboard of KITTI Vision Benchmark between 2014 and 2021. 
    He founded Kaarta, Inc, a CMU spin-off commercializing 3D mapping \& modeling technologies, and stayed with the company for 4 years as chief scientist.
    His research interests are in robotic navigation, spanning localization, mapping, planning, and exploration.
\end{IEEEbiography}

\begin{IEEEbiography}[{\includegraphics[width=1in,height=1.25in,clip,keepaspectratio]{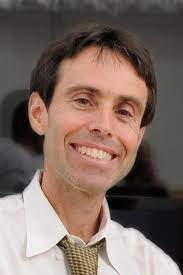}}]
    {Howie Choset} received the B.S. Eng. degree in computer science and the B.S. Econ. degree in entrepreneurial management from the University of Pennsylvania (Wharton), Philadelphia, PA, USA, both in 1990, the M.S. and Ph.D. degrees in mechanical engineering from California Institute of Technology (Caltech), Pasadena, CA, USA, in 1991 and 1996, respectively.
    He is currently a Professor of Robotics with the Carnegie Mellon University, Pittsburgh, PA, USA. His research group reduces complicated high dimensional problems found in robotics to low-dimensional simpler ones for design, analysis, and planning. 
\end{IEEEbiography}

\vspace{-1cm}
\begin{IEEEbiography}[{\includegraphics[width=1in,height=1.25in,clip,keepaspectratio]{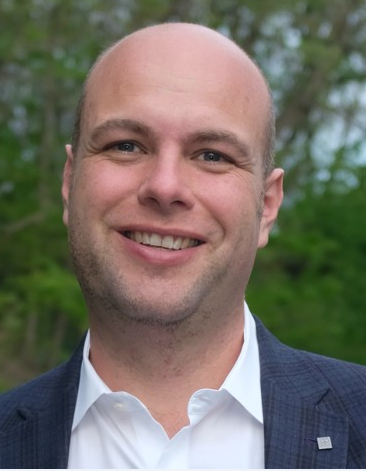}}]
    {Sebastian Scherer} received his B.S. in Computer Science, M.S. and Ph.D. in Robotics from CMU in 2004, 2007, and 2010. 
    Sebastian Scherer is an Associate Research Professor at the Robotics Institute at Carnegie Mellon University. His research focuses on enabling autonomy for unmanned rotorcraft to operate at low altitude in cluttered environments. He is a Siebel scholar and a recipient of multiple paper awards and nominations, including AIAA@Infotech 2010 and FSR 2013. His research has been covered by the national and internal press including IEEE Spectrum, the New Scientist, Wired, der Spiegel, and the WSJ. 
\end{IEEEbiography}

\end{document}